\documentclass[preprint,12pt]{elsarticle}

\usepackage{amssymb}

\usepackage{amsmath}
\usepackage{multirow}
\usepackage{subcaption}
\usepackage{booktabs} 

\newcommand{\norm}[1]{\left\lVert#1\right\rVert}

\DeclareMathOperator*{\argmax}{arg\,max}

\journal{Neural Networks}

\begin{document}

\begin{frontmatter}



\title{Adaptive Temperature Scaling for Robust Calibration of Deep Neural Networks}


\author[AUDIAS]{Sergio A.Balanya\corref{mycorrespondingauthor}\fnref{fn1}}
\cortext[mycorrespondingauthor]{Corresponding author.}
\ead[url]{http://audias.ii.uam.es}
\ead{sergio.alvarezb@estudiante.uam.es}
\fntext[fn1]{This work was written prior to Sergio A.Balanya joining Amazon.}
\author[MLgroup]{Juan Maroñas}
\author[AUDIAS]{Daniel Ramos}

\address[AUDIAS]{AUDIAS Laboratory - Audio, Data Intelligence and Speech. \\ Escuela Politecnica Superior. Universidad Autónoma de Madrid. \\ Calle Francisco Tomás y Valiente 11. 28049 Madrid. Spain.}

\address[MLgroup]{Machine Learning Group. \\ Escuela Politecnica Superior. Universidad Autónoma de Madrid. \\ Calle Francisco Tomás y Valiente 11. 28049 Madrid. Spain.}

\begin{abstract}
In this paper, we study the post-hoc calibration of modern neural networks, a problem that has drawn a lot of attention in recent years. Many calibration methods of varying complexity have been proposed for the task, but there is no consensus about how expressive these should be. We focus on the task of confidence scaling, specifically on post-hoc methods that generalize Temperature Scaling, we call these the Adaptive Temperature Scaling family. We analyse expressive functions that improve calibration and propose interpretable methods. We show that when there is plenty of data complex models like neural networks yield better performance, but are prone to fail when the amount of data is limited, a common situation in certain post-hoc calibration applications like medical diagnosis. We study the functions that expressive methods learn under ideal conditions and design simpler methods but with a strong inductive bias towards these well-performing functions. Concretely, we propose Entropy-based Temperature Scaling, a simple method that scales the confidence of a prediction according to its entropy. Results show that our method obtains \textit{state-of-the-art} performance when compared to others and, unlike complex models, it is robust against data scarcity. Moreover, our proposed model enables a deeper interpretation of the calibration process.
\end{abstract}


\begin{keyword}
 Robust Calibration \sep Deep Neural Networks \sep Over-confidence \sep Uncertainty \sep Classification 
\end{keyword}

\end{frontmatter}


\section{Introduction}\label{sec:intro}

There is an increasing trend in using Deep Neural Networks (DNNs) to automate a multitude of tasks, like image classification for healthcare \cite{health} and speech recognition \cite{speech} among others. Some of these are high-risk applications, for example, a False Negative in cancer detection could be fatal for the patient. To this end, it is of paramount importance to use reliable Machine Learning (ML) systems that acknowledge the uncertainty of their predictions. A probabilistic classifier that outputs a confidence value, or probability, for each class, allows to make Bayes decisions---i.e. optimum decisions leveraging the cost of such decisions \cite{dudayhart}. 

The extent to which the confidence outputs of a classifier can be interpreted as class probabilities is what is known as the \textit{calibration} of a classifier \cite{dawid82, degroot83}. Modern DNNs achieve very low test error rates but are not necessarily well-calibrated \cite{Guo2017, Ovadia2019}. Hence, the focus of the community is shifting towards improving the calibration of DNNs.

One approach to obtain better confidence estimates is to average the predictions of different models using ensembles \cite{Lakshminarayanan2017} or taking a Bayesian approach \cite{Kristiadi2020}. Data Augmentation techniques have also been used to improve calibration \cite{Patel2020b, augmix}, as well as modified training objectives \cite{softCalib, PTS}. Among popular approaches, and the focus of this work, is the approach of \textit{post-hoc} calibration, in which the predictions of an already trained classifier are re-calibrated. Typically, a new model, the calibrator, is trained on the outputs of the classifier evaluated on a held-out dataset. This approach results very convenient since one can use off-the-shelf ML systems that already present good test error rates taking advantage of a plethora of work on DNNs. Deep Learning models have been widely adopted and usually offer a good solution for any Machine Learning task. For this reason, DNNs have become standard models with an easy application via public frameworks like Pytorch \cite{pytorch} and Tensorflow \cite{tensorflow}. With post-hoc calibration, we may still apply typical DNNs to high-risk tasks and benefit from their good error rates without over-confidence issues.

Probably the most popular post-hoc calibration method is Temperature Scaling (TS) proposed by \cite{Guo2017}. It is a single parameter model that re-scales the confidence predictions by a \textit{temperature factor} for its use with DNNs. The simplicity of this method and the fact that in their experiments it seems to perform better than more complex ones lead authors to believe that the problem of re-calibration is inherently simple. However, recent alternatives based on expressive models like Bayesian Neural Networks \cite{Maronas2019} and Gaussian Processes \cite{Wenger2020} improve TS, suggesting that re-calibration might be a more complex problem than it was previously assumed. However, expressive models can be more data-hungry and may require careful tuning when the amount of data is limited.

Based on the observation that miscalibration on modern DNNs is often caused by over-confidence \cite{Guo2017, Minderer2021}, recent work proposes to learn more complex calibration functions than TS but from a constrained space. By imposing some restrictions, like being Accuracy-preserving \cite{Zhang2020} and order-invariant \cite{intraOrder}, authors force an inductive bias towards the desired calibration functions. This approach shows promising results, but it may still fail in low-data scenarios, especially when using over-parametrized models. This can be a huge limitation in tasks where data for calibration is usually scarce: In certain language recognition tasks \cite{nistlre17} some languages can be underrepresented; also, it can be difficult to obtain training examples for the medical diagnosis of very rare diseases \cite{Yoo2021}. Hence, there is a need for calibration methods that achieve high performance with low data requirements.

To this end, choosing a model with a suitable inductive bias gives some advantages. First, the set of possible calibration functions that the model can learn, or hypothesis space, is reduced. This translates into an easier training objective. Moreover, if the bias is well-specified, the learned calibration function will be more robust against a lack of training data, and will better generalize to other data \cite{inductiveBias}. The quality of the inductive bias depends on the knowledge we have of the task at hand. For instance, the specific architecture of Convolutional Neural Networks (CNNs), based on convolution filters, explains their success on visual recognition tasks \cite{lenet5}, even though by sharing weights the total number of parameters is reduced, thus limiting the learning capacity.

\subsection{Contributions}

Intending to gain knowledge about the specific task of modern DNNs calibration, we provide a study of post-hoc adaptive calibration methods, with varying degrees of expressiveness and robustness, that lead to better calibration. This may help design models more resilient to data scarcity. We focus on the problem of confidence scaling as the bad-calibration properties of DNNs are mainly attributed to over-confidence \cite{Guo2017}. 

To perform this study we focus on Adaptive Temperature Scaling (ATS) methods, a family of calibration maps that generalizes TS by making the \textit{temperature factor} input-dependent as proposed by \citet{localTS}. However, the authors propose to estimate the temperature factor as a function of the classifier input. ATS models, on the other hand, learn a \textit{temperature function}, that computes temperature factors directly from the output of the classifier. Within this family, we can compare several calibration methods which extend the expressiveness of TS in different ways.

We analyze and benchmark several calibration models focusing on which temperature functions can lead to better calibration. Results show that highly parametrized methods achieve high performance when there is plenty of data, but also that these are doomed to failure in low-data scenarios. By exploiting gained knowledge about the post-hoc calibration task, we develop Entropy-based Temperature Scaling (HTS), a method with a strong inductive bias that is robust to the size of the dataset and provides comparable performance to other \textit{state-of-the-art} methods.

The rest of the paper is organized as follows. First, we introduce some theoretical background of the calibration task. Then, in Section \ref{sec:meth} we introduce some post-hoc calibration methods and motivate their design. We also describe other existing techniques to which we compare our methods. In Section \ref{sec:exp} we describe the performed experiments and show their results. Finally, in the last section, we give our conclusions and comment on possible future work.


\section{Background}

In this work we focus on the multi-class classification task. Let $x \sim X \in \mathcal{X}$ be the input random variable with associated target $y \sim Y \in \mathcal{Y}$, where $y = [y_1, y_2, .., y_K]$ is a one-hot encoded label. The goal is to obtain a probabilistic model $f$ for the conditional distribution $P(Y|X=x)$. The model defines the function $f(x) = z, \, x \in \mathcal{X}, \, z \in \mathbb{R}^K$. The outputs $z$ of the model are known as \textit{logits} since they are later mapped to probability vectors via the softmax function:
\begin{equation*}
    q = \sigma_{SM}(z) = \frac{\exp z}{\sum_{k=1}^K \exp z_k},
\end{equation*}
where the exponential in the numerator is applied element-wise, and $q \in \mathbb{S}^K$ is the corresponding probability vector. We use $\mathbb{S}^K$ to denote the probability simplex in $K$ classes.

In practice there is no distribution $P(X, Y)$ (or we do not have access to it). Instead, we have a labeled data set $\mathcal{D}$ of $N$ pair-realizations $\mathcal{D} = \{x^{(i)}, y^{(i)}\}_{i=1}^N$ that is used to approximate it. For example, DNNs are normally trained by minimizing the expected value of some cost function. This expected value is computed from the empirical distribution induced by placing a Dirac's \textit{delta} at each point $\{x^{(i)}, y^{(i)}\}_{i=1}^N$.

\subsection{Calibration}

A probabilistic classifier is said to be \textit{well-calibrated} whenever its confidence predictions for a given class match the chances of that class being the correct one \cite{dawid82, degroot83}. We can express this property as an equation in terms of the probability distributions introduced earlier:
\begin{equation}
    P(y\,|\,q) = q, \,\, \forall q \in \mathbb{S}^K,
\end{equation}
where $P(y\,|\,q)$ represents the relative class frequency ---i.e. the proportion of each class on the set of all samples for which the classifier predicts $q$.

From this expression, it is easy to derive a measure of miscalibration or \textit{Calibration Error} (CE):
\begin{equation}
    CE = \mathbb{E}_{P(X, Y)}\big[ \norm{P(y\,|\,q) - q}_d\big].
\end{equation}

This is, the expected value of the $d$-norm of the difference between prediction vectors and the relative class proportions.

While this equation might be useful to illustrate the concept of miscalibration, it does not provide a feasible way to measure it. Our main problem is the non-existent $P(X, Y)$. First, we cannot compute the expected value w.r.t. a non-existent distribution. Yet the main limitation is evaluating the distribution since one can use the empirical distribution given by the labelled set $\mathcal{D}$ as MC samples. However, there is no simple way of evaluating $P(y\,|\,q)$ using this empirical distribution. Therefore, further approximations are required to estimate the miscalibration of a classifier.

\subsubsection{ECE}

The most popular metric used to estimate the Calibration Error is the \textit{Expected Calibration Error} (ECE) \cite{Guo2017, Naeini2015}. This metric uses a histogram approach to model $P(y\,|\,q)$ and considers only top-label predictions. The samples of a given evaluation set $\mathcal{D}_{test}$ are partitioned into $M$ bins ${B_1, B_2, ..., B_M}$ according to the confidence of their top prediction:
\begin{equation*}
    B_i := \Big\{(x, y) \in \mathcal{D}_{test} \;:\; \frac{i-1}{M} < \max (q) \leq \frac{i}{M} \Big\}
\end{equation*}

Then the ECE is computed as:
\begin{equation}
    ECE =  \sum_{i=1}^M \frac{|B_i|}{|\mathcal{D}_{test}|}\,|acc(B_i) - conf(B_i)|,
\end{equation}\label{eq:ECE}
where $|\cdot|$ denotes the number of samples in a set, $acc(B_i)$ is the accuracy of the classifier evaluated only on $B_i$, and $conf(B_i)$ is the mean confidence of the top-label predictions in $B_i$.

Despite its popularity, this estimator provides unreliable results as it is biased and noisy \cite{Verified_Uncertainty_Calibration, Vaicenavicius2020, Nixon2019}. Many improvements over the ECE have been proposed to mitigate these problems such as class-wise ECE and using variable confidence intervals \cite{Nixon2019}. However, there is not any binning scheme consistently reliable \cite{Simonoff1997}. Anyway, ECE remains the most popular metric used by the community to measure miscalibration and we use it in our experiments to report results for the sake of comparison.

\subsubsection{Proper Scoring Rules}

One way to implicitly measure calibration is to use Proper Scoring Rules (PSRs). Any PSR can be decomposed into the sum of two terms \cite{Brocker09}, a \textit{refinement} term and the so-called \textit{reliability} or \textit{calibration}. Thus, when evaluating the goodness of a classifier with a Proper Scoring Rule, one is also indirectly measuring calibration. The fact that the calibration component cannot be evaluated in isolation is what drives the community to use approximated metrics like ECE. Moreover, different PSRs may rank differently the same set of systems evaluated on the same data. Nevertheless, PSRs provide a theoretically grounded way of measuring the goodness of a classifier. Throughout this work, we will use two different PSRs to evaluate models, the log-score or Negative Log-Likelihood (NLL) and the Brier score, both of them well-known \cite{Dawid2014}.

\subsection{Post-hoc Calibration}

Ideally, a model $f$ trained on some data $\mathcal{D}$ would \textit{generalize} and show good calibration properties when evaluated on other data $\mathcal{D}_{test}$, assuming both sets are reasonably similar. However, many classification systems turn out to be badly calibrated in practice, for instance, Convolutional Neural Networks (CNNs) tend to produce overconfident predictions \cite{Guo2017, Minderer2021}. Moreover, in some tasks, it cannot be guaranteed that the training data is similar enough to the actual data on which the model will be deployed. For instance, a language recognition system may be trained on broadcast narrowband speech (BNBS) data but applied in a telephone service where the audio characteristics are different. To solve this problem, one common approach is that of post-hoc calibration, in which a function is applied to the outputs of the model. This function can be seen as a decoupled classifier that learns to map uncalibrated outputs to calibrated ones---i.e. $q \mapsto \hat{q}$. We use the $\hat{\cdot}$ notation to denote the \textit{calibrated} prediction. The standard practice is to fit this calibration map or calibrator in a held-out data set $\mathcal{D}_{val}$, or validation data, that is supposed to resemble the data on which the model will make predictions.

Many post-hoc calibration methods take as input prediction \textit{logits} instead of the final probability vectors. Notice that this does not limit their applicability since the outputs $q$ of a probabilistic model can be mapped to the logit domain through the logarithmic function $z = \log q + k$, where $k$ is an arbitrary scalar value.

\subsubsection{Accuracy-preserving Calibration}

Modern classification systems achieve very low test error rates and their miscalibration is attributed mainly to over-confidence---i.e. predicted confidences that call for higher accuracy rates than those actually obtained. Under this assumption, it is reasonable to constrain the calibration transforms so that the predicted ranking over the classes is maintained. This condition is known as Accuracy-preserving \cite{Zhang2020} because functions that meet it do not change the top-label prediction, $\argmax q = \argmax \hat{q}$.

When using expressive, and unconstrained, classification models like DNNs for the task of calibration, it is possible to improve calibration at the cost of losing accuracy \cite{Maronas2019, intraOrder}. This trade-off is avoided by restricting the calibration functions to be Accuracy-preserving so that the class decision, left to the classifier, is decoupled from the confidence estimation of each decision. 

In this work, we compare only Accuracy-preserving methods and avoid a potential problem often encountered in the calibration task. Since miscalibration is measured in isolation, accuracy is also considered to evaluate calibrators. This poses the question of determining which calibrator is better, one that improves more calibration but degrades the accuracy, or one that does not degrade the accuracy but shows less improvement on calibration. This decision is often application dependent but can be circumvented by using an Accuracy-preserving method.

\subsubsection{Temperature Scaling} 

Temperature Scaling (TS) is probably the most widely used post-hoc calibration approach in the literature. It belongs to the family of Accuracy-preserving methods. It scales the output logits by a \textit{temperature factor} $T_0\in \mathbb{R}^+$:
\begin{equation}
    \hat{z} = \frac{z}{T_0}.
\end{equation}

This factor is obtained by minimizing the NLL on some validation data consisting of predictions of the uncalibrated classifier. Since the NLL is a Proper Scoring Rule, TS is encouraged to improve calibration. Consequently, the temperature factor conveys information about the level of over-confidence in these predictions. A high temperature $T_0 > 1$ flattens the logits so the probability vectors approach the uniform distribution $q = [1/K, 1/K, ..., 1/K]$, thus relaxing the confidences and fixing over-confidence. On the other hand, a low temperature $T_0 < 1$ sharpens the confidence values moving the top-label predictions towards $1$ and the others towards $0$. Hence, fixing under-confidence.


\section{Methods}\label{sec:meth}

In this section, we first describe the Adaptive Temperature Scaling family and illustrate it by proposing some methods of our contribution. Then, we introduce other Accuracy-preserving methods, not necessarily of the ATS family, with state-of-the-art performance that we use as benchmarks in the experiments.

\subsection{ The Adaptive Temperature Scaling family}

We refer to the group of Accuracy-preserving maps that generalizes Temperature Scaling as the ATS family. All ATS methods can be expressed as the calibration function:
\begin{equation}
    \hat{z} = \frac{z}{T(z)},
\end{equation}\label{eq:ATS}
where $T: \mathbb{R}^K \mapsto \mathbb{R}^+$ is the \textit{temperature function}.

This family generalizes Temperature Scaling by making the temperature factor input-dependent. TS is limited to the temperature function $T(z) = T_0$, where $T_0$ is the scalar parameter of the model. Hence, TS implicitly assumes that a classifier will generate predictions with the same level of over-confidence independently of the specific sample being classified.

On the other hand, a general ATS method computes a different temperature factor for each prediction via the temperature function $T(z)$. The computed factor for some $z$ estimates the degree of over-confidence of the corresponding prediction $q = \sigma_{SM}(z)$. Hence, ATS methods acknowledge the possibility that a classifier's over-confidence may depend on the samples being classified.

The input-dependent property was first exploited by \citet{localTS} with their Local Temperature Scaling method. However, this approach relies on the classifier input $x$ to estimate a temperature factor $T_x = T(x)$. An ATS method estimates the factor based on the classifier output instead, $T_x = T(z)$, thus separating further the calibration step from the original classification task. The former approach tries to learn for which inputs the classifier is likely overconfident. ATS is independent of the classification task and is only concerned with estimating the overconfidence of an already made prediction. In other words, Local TS should be tailored for each classification task, for instance, if the input is audio one might use an RNN but choose a CNN instead for images. On the other hand, the input space of ATS methods is always the logit domain so these are more likely to generalize across classification tasks.

We acknowledge that this may reduce the potential expressive power of ATS since $z$ is a processed version of $x$. Nevertheless, we believe that such constraint is not necessarily limiting since, as we show in our experiments, the logit vector of a prediction already conveys information about its degree of miscalibration. Moreover, one advantage of post-hoc methods is the decoupling of the classification step from the calibration step. This is in some sense lost if the original classifier input is required for the calibration.

\subsection{Proposed Methods}

We introduce three different ATS methods based on simple temperature functions. These functions are theoretically motivated and interpretable, so we can empirically validate the use of more expressive calibration transforms. First, we note that to meet the positivity constraint on the temperature factor we apply the softplus function to our models' outputs:

\begin{equation}
    \sigma_{SP}(a) = \ln (1 + e^a).
\end{equation}

\subsubsection{Linear Temperature Scaling} 

We call this method Linear Temperature Scaling (LTS) since it is based on a linear combination of the logit vector, its temperature function is given by:

\begin{equation}
    T_{LTS}(z) = \sigma_{SP}(w^{L}\cdot z + b),
\end{equation}
where $w^{L} \in \mathbb{R}^K$ and $b \in \mathbb{R}$ are the learnable parameters of the model. 

The weight vector $w^L$ takes into account the score assigned to each class to determine the level of over-confidence. Hence, LTS can predict higher temperature factors for certain predicted classes than for others. The scalar parameter $b$ allows LTS to recover the base TS by zeroing the $w^L$ parameter.

We motivate this method by giving the following example: an uncalibrated classifier can make over-confidence predictions for only certain classes. Since LTS weights each component of the logit vector to obtain the temperature factor, it should be able to raise (shrink) it by increasing (decreasing) the weight component $w_i^L$ depending on whether the classifier is more (less) likely to make and over-confident prediction when predicting class $i$.

From this follows the interpretation of the method. After fitting LTS on a validation set, the weight vector will point towards the direction of the highest degree of overconfidence in the logit space.

\subsubsection{Entropy-based Temperature Scaling} 

Motivated by the fact that the entropy of the predictive class distribution can be interpreted as the uncertainty of such prediction, we propose HTS. The temperature function of this method is given by:
\begin{equation}
    T_{HTS}(z) = \sigma_{SP}\left(w^H\log \overline{H}(z)  + b \right),
\end{equation}
where $\overline{H}(z) = H(\sigma_{SM}(z))/\log K$ is the \textit{normalized entropy}, and $w^H \in \mathbb{R}$ and $b \in \mathbb{R}$ are the learnable parameters of the model. We normalize the entropy so that it is always upper-bounded by $1$ irrespective of the number of classes. This allows us to generalize the interpretation of $w^H$ between tasks with a different number of classes. We apply the logarithm to the entropy because, as we show later in the experiments, the temperature shows a linear trend with the logarithm of the entropy. We give $b$ the same interpretation as in the previous model. The parameter $w^H$ determines how much the predictive uncertainty of predictions---i.e. the $\log \overline{H}(z)$---influence the determination of the temperature factor. The higher the magnitude of $w^H$ the more variability we can expect in the computed temperature factors. On the other hand, a model with $w^H \to 0$ will resemble the base TS.

The ECE metric and over-confidence evaluation \cite{Minderer2021}, are tasks that consider only the confidence value assigned to the top-rated class. This value represents the class probability estimated by the classifier. While it is a confidence value, it does not represent the `confidence' of the classifier on the prediction, it just concerns the predicted class in particular. Conversely, the entropy of the predictive is a measure of uncertainty of the whole prediction---i.e. an alternative more comprehensive way of assessing the classifier's confidence in some prediction.

For instance, we may have two predictions $q^{(i)} = [0.6, 0.2, 0.2]$ and $q^{(j)} = [0.6, 0.4, 0.0]$ in a 3-class problem. Both assign the same confidence $0.6$ to class 1, but it is clear that $q^{(i)}$ is a higher entropy predictive than $q^{(j)}$---i.e. it is a more uncertain prediction. 

Again, we motivate the method with a hypothetical example. Suppose that we have a classifier that produces predictions with variable degrees of over-confidence. One way in which a prediction-logit can convey information about its level of over-confidence is via its entropy. This is, for two predictions with the same predicted confidence, we may assume that the more uncertain of the 2---i.e. the higher entropy prediction---is more likely to be over-confident since it reports the same value of confidence despite its higher uncertainty. 

This model makes a strong assumption about the level of over-confidence in a prediction. Mainly, that it can be expressed as a simple linear function of the log-entropy. The resulting model is easy to train since the set of possible calibration functions, or hypothesis space, is comparatively limited. However, its performance is completely conditioned on the assumption being met. We provide experiments validating the model in Section \ref{sec:exp}.

\subsubsection{Combined system} 

Finally we propose HnLTS, a model that combines the previous two with a single temperature function given by:
\begin{equation}
    T_{HnLTS}(z) = \sigma_{SP}\left(w^{L}\cdot z + w^H\log \overline{H}(z)  + b \right),
\end{equation}
where $w^{L} \in \mathbb{R}^K$, $w^H \in \mathbb{R}$, and $b \in \mathbb{R}$ are the learnable parameters to which we give the same interpretation as above. 

The motivation behind this model is to increase the expressiveness of the system in a controlled way to see how this affects its performance and training procedure compared to the simpler methods. The hypothesis space of this method is a combination of the previous two so it should be able to recover the solution of either one. However, we argue that the increased hypothesis space also makes the model more difficult to train with higher data requirements.

\subsection{Baseline Methods}\label{sec:base}

We now describe other Accuracy-preserving methods already existing in the literature with \textit{state-of-the-art} performance. Some of these, but not all of them, belong to the ATS family as they can be expressed in the general form given by Equation \ref{eq:ATS}. 


\subsubsection{Parametrized Temperature Scaling}

Parametrized Temperature Scaling (PTS) \cite{PTS} is a specific instance of the ATS family in which the temperature function is conditioned to be a neural network (NN). The input to the NN is the logit vector sorted by decreasing value of confidence $z^s$. Sorting the logit vector makes the model order-invariant \cite{intraOrder} simplifying the hypothesis space at the cost of losing the possibility of discriminating between classes---i.e. it cannot consider the predicted ranking over the classes. PTS can be expressed as an ATS method with temperature function:
\begin{equation}
    T_{PTS}(z) = \textit{NN}(z^s),    
\end{equation}
where \textit{NN} is the function defined by the neural network.

Instead of optimizing the parameters of the NN to minimize some PSR as other methods do, authors propose to minimize an ECE-based loss given by:
\begin{equation}
    L_{ECE} = \sum_{i=1}^M \frac{|B_i|}{|\mathcal{D}_{test}|}\,\norm{acc(B_i) - conf(B_i)}_2,
\end{equation}\label{eq:L_ECE}
where $B_i$, $conf(B_i)$, and $acc(B_i)$, are defined as in Equation \ref{eq:ECE}. During training, samples are re-partitioned into ${B_i}$ at each loss evaluation since the confidence is re-scaled differently.

In their experiments, authors always use the same architecture, a Multi-Layer Perceptron (MLP) with two 5-unit hidden layers. Authors limit the input size of the network to the 10 highest confidence values whenever the number of classes is greater than 10. We use the same architecture in our experiments.

\subsubsection{Bin-Wise Temperature Scaling}

Bin-Wise Temperature Scaling (BTS) \cite{bts} is a histogram-based method that applies a different temperature factor to each bin of the histogram. First, test samples are partitioned into $N$ bins according to their top-label confidence. Authors force a high-confidence bin that ranges from 0.999 to 1. The samples with predicted confidence below 0.999 are partitioned into the other $N-1$ intervals such that each bin contains the same number of samples.

This method can also be included in the ATS family. The temperature function in this case is just a look-up table that assigns the corresponding temperature factor to the input confidence value.

\subsubsection{Ensemble Temperature Scaling}

Ensemble Temperature Scaling (ETS) \cite{Zhang2020} obtains a new logit vector as a convex combination of the uncalibrated vector, a maximum entropy logit vector, and the temperature-scaled vector:
\begin{align}
    \hat{z} = w_1\frac{z}{T_{ETS}} + w_2 z + w_3 \frac{1}{K}, \\
    \text{subject to}
    \, w_1+w_2+w_3 = 1; w_i \geq 0 \notag
\end{align}
where $w_1$, $w_2$, $w_3$ are the learnable weights of the convex combination and $T_{ETS}$ is the temperature parameter of the TS component. All the parameters are optimized en bloc to minimize some PSR.

This method is Accuracy-preserving and also an extension of the standard TS, however, it does not belong to the ATS family. This can be easily verified by noting that ATS methods compute for some logit vector a single scalar temperature factor which applies equally to every entry of the logit vector. On the other hand, ETS scales by a different temperature factor each component of the logit vector.




\section{Experiments}\label{sec:exp}

We present two rounds of experiments. First, we report a study of the proposed methods that motivate their design and present ways in which calibration performance improves with model complexity. With these, we give evidence that the logit vector conveys information about its degree of over-confidence and motivates the design of new calibration methods that takes this into account. Then, we compare our methods with other state-of-the-art Accuracy-preserving calibration techniques in different dataset-size settings to assess their robustness to data scarcity.

\begin{figure}
	\centering {
	\includegraphics[width=0.8\linewidth]{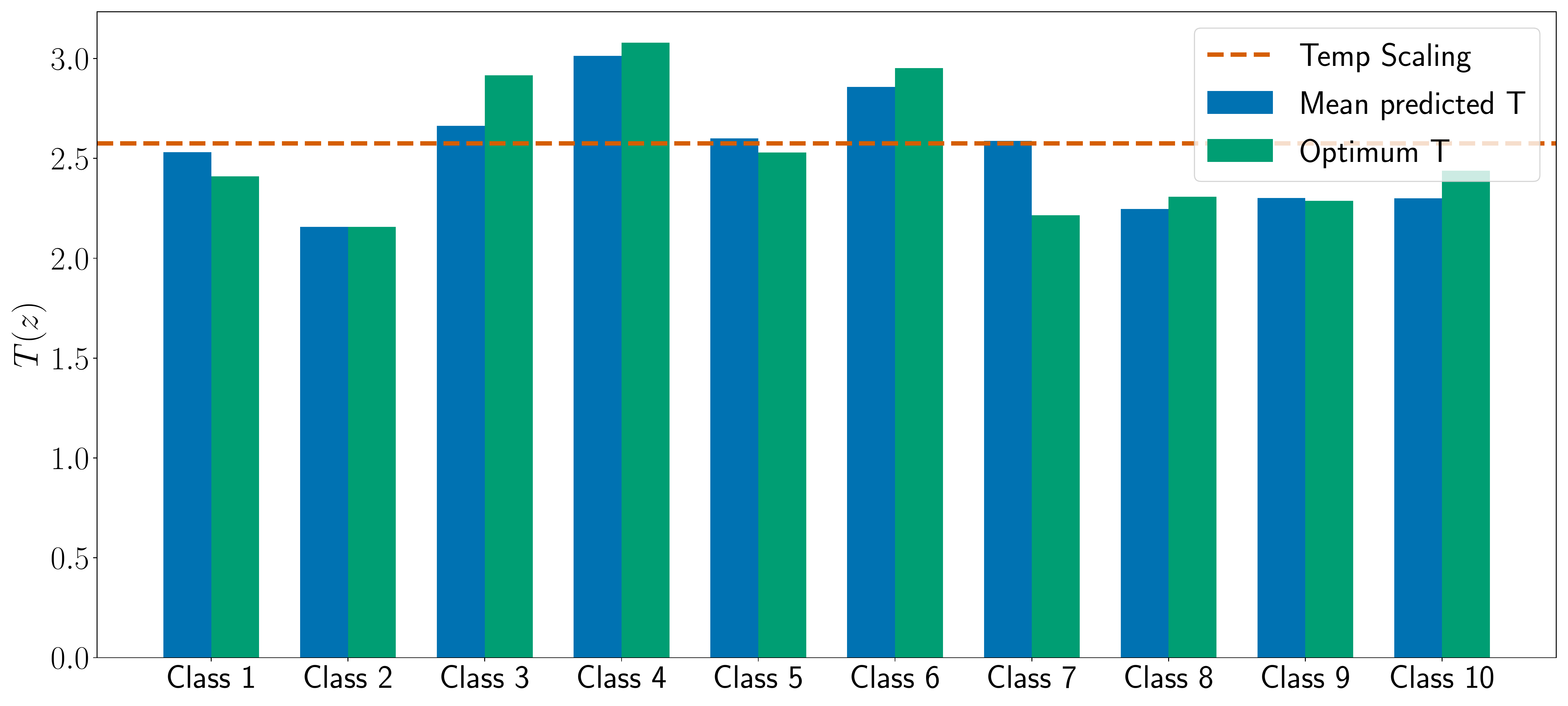}
	}
	\caption{Mean predicted Temperature (blue) against optimum temperature factor (green) for each class on the test set.}\label{fig:class_LTS}
\end{figure}

\subsection{Setup}

\subsubsection{Datasets and tasks}

We refer to \textit{model-dataset} pairs as calibration tasks. So a task is composed of the predictions of a \textit{model}, for instance a ResNet-101 \citep{resnet}, on a specific \textit{dataset}, like CIFAR-100 \citep{CIFAR}. Every dataset is partitioned into three splits: train, validation, and test. The model of each task is trained using the train set and then it is used to generate predictions on the validation and test sets. We evaluate a calibration method on a certain task using the following procedure: First we fit the calibration method using the predictions on the validation set. Then we apply it to the test set predictions and compute metrics over these.

\subsubsection{Training details}

We use NLL as the optimization objective to fit calibrators. Additionally, in all tasks, we fit a second version of the PTS method, minimizing the ECE-based loss instead (see Section \ref{sec:base}). All methods except TS and ETS are implemented in Pytorch \citep{pytorch} and optimized using Stochastic Gradient Descent (SGD) with an initial learning rate of $10^{-4}$, Nesterov momentum \citep{nesterov} of $0.9$, and a batch size of 1000. We reduce the learning rate on plateau by a factor of 10 until the learning rate reaches $10^{-7}$ that we stop training considering the algorithm has converged. The standard TS is optimized with SciPy \citep{scipy} and the Broyden–Fletcher–Goldfarb–Shanno (BFGS) algorithm. To calibrate with ETS we use the code uploaded by authors \cite{Zhang2020}. 

All experiments are run 50 times with different random initializations and the results are averaged across runs. For the experiments in which a subsampled validation set is used, this is resampled at each run but consistent across calibration methods. This is, for each of the 50 runs we sample a $N$-sized validation set and use it to fit all the calibration methods in the comparison.


\subsection{Analysis of the ATS methods}

For the first round of experiments, we calibrate a ResNet-50 \cite{resnet} on CIFAR-10 \cite{CIFAR} with the proposed interpretable ATS methods, Entropy-based TS (HTS) and Linear TS (LTS), and discuss each separately.

\subsubsection{Linear TS: Introducing class dependence}

With this experiment, we aim to illustrate the example that we give to motivate the LTS method. This is, that LTS can adapt to a classifier that makes more o less over-confident predictions depending on which class it predicts as correct.

We divide the test set of predictions according to their true class and compute for each subset the optimum temperature factor, which is obtained by optimizing TS on each group. Then, we use the LTS model optimized in the validation set to compute a temperature factor for every test prediction. Finally, we represent in Figure \ref{fig:class_LTS} the average of these factors per subset against the optimum temperature. For reference, we include the TS temperature factor learned on the validation set (dashed orange line).

From Figure \ref{fig:class_LTS} we notice that the classifier does produce more over-confident predictions for some classes than for others, even in a curated and well-balanced dataset such as CIFAR-10. We can expect this effect to be even more present in real-life applications in which the prevalence of classes may vary and some distribution mismatch between development and production data can be expected. LTS exploits this difference between classes and manages to adapt the temperature factor in each subset closely matching the optimum.

\subsubsection{Entropy-based TS: Leveraging uncertainty of predictions}

Our motivation for the HTS method is that the level of over-confidence in a prediction is related to the entropy of such prediction. If our hypothesis is correct, we can expect, for the same value of confidence in the predicted class, higher entropy predictions like q(i) to be more over-confident on average. So, we might expect higher temperature factors for higher entropy predictions.

\begin{figure}
     \centering
     \includegraphics[width=\linewidth]{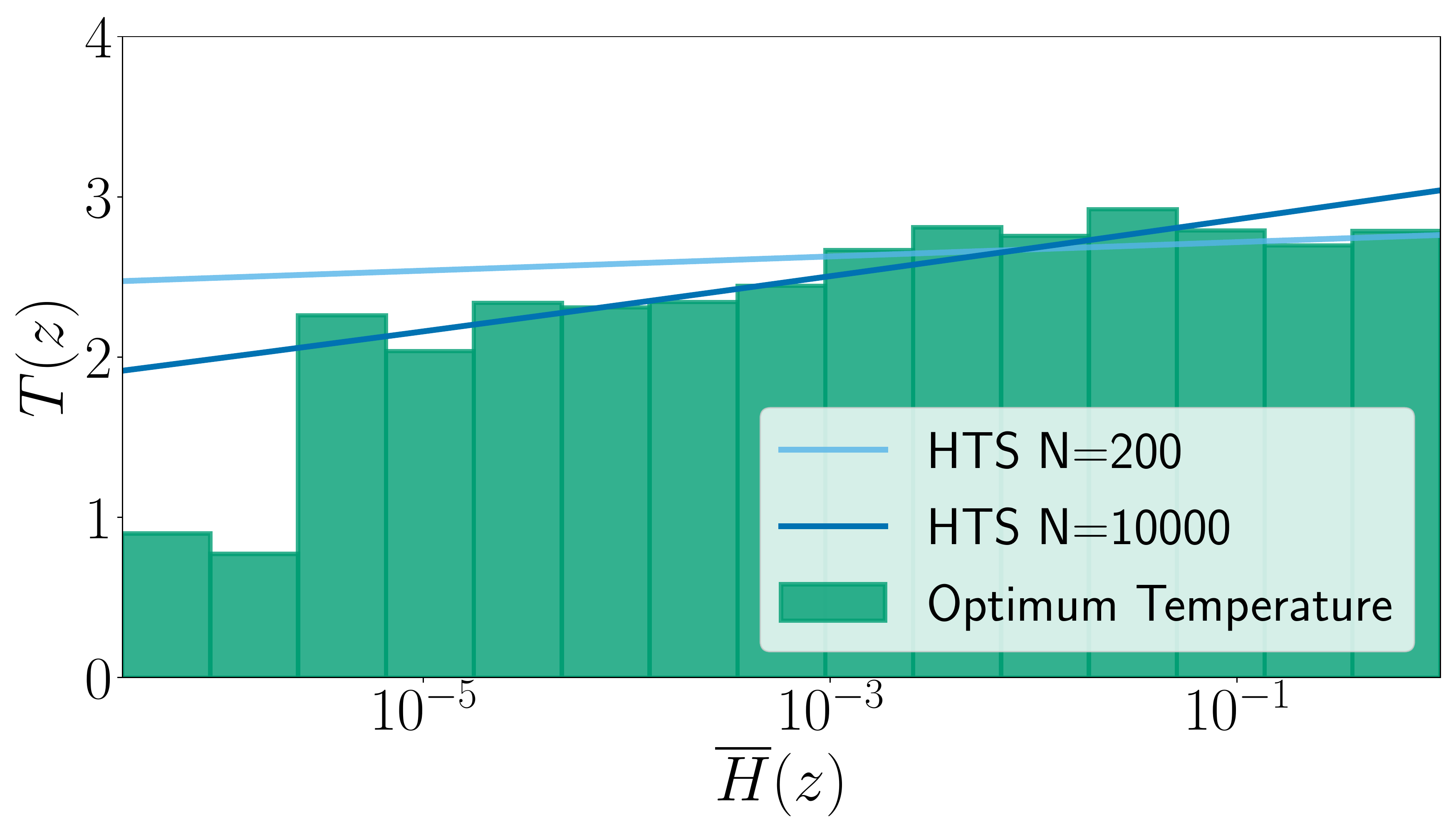}
     \caption{Temperature function of HTS fitted using 200 (light blue) and 10000 (dark blue) validation samples and optimal temperature on the test set (green).}
     \label{fig:h_HTS}
 \end{figure}

 \begin{figure}
     \centering
     \includegraphics[width=\linewidth]{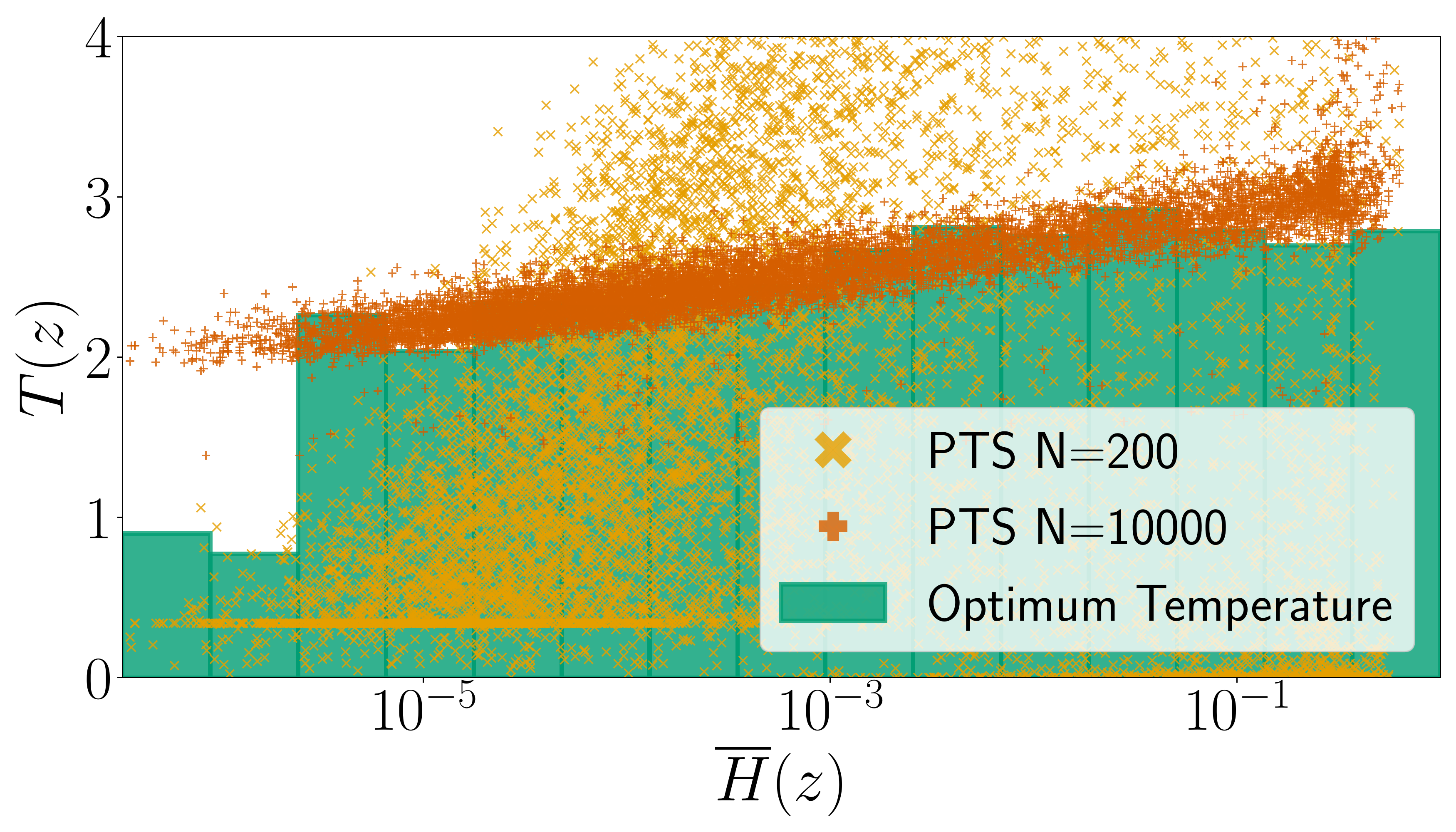}
     \caption{Temperature factors of PTS for test samples fitted using 200 (light orange) and 10000 (dark orange) validation samples and optimal temperature on the test set (green).}
     \label{fig:h_PTS}
 \end{figure}

In Figure \ref{fig:h_HTS} we depict the temperature function learned by HTS in the validation set. We train two models, one with the full validation set, plotted in a darker shade, and the other using a random subset of 200 samples. We also plot the optimum temperature factor estimated in the test set for different ranges of normalized entropy. We partition the log-domain of the normalized entropy in equally spaced bins and divide the test samples according to this binning scheme. For each bin, we estimate the optimum temperature factor given by TS. In a second experiment, we show the temperature factor that PTS assigns to each prediction on the test set (see Figure \ref{fig:h_PTS}). With this plot, we aim to see if a very expressive method like PTS learns any relation between the entropy of a prediction and its temperature factor.

We find that, at least in this particular task, there exists some positive relation between the entropy of the predictive and its level of over-confidence. Figure \ref{fig:h_HTS} shows that a linear function is a fair approximation to the relation between entropy and temperature and that HTS manages to capture it even in the face of low data.

 \begin{figure}
     \centering
     \includegraphics[width=\linewidth]{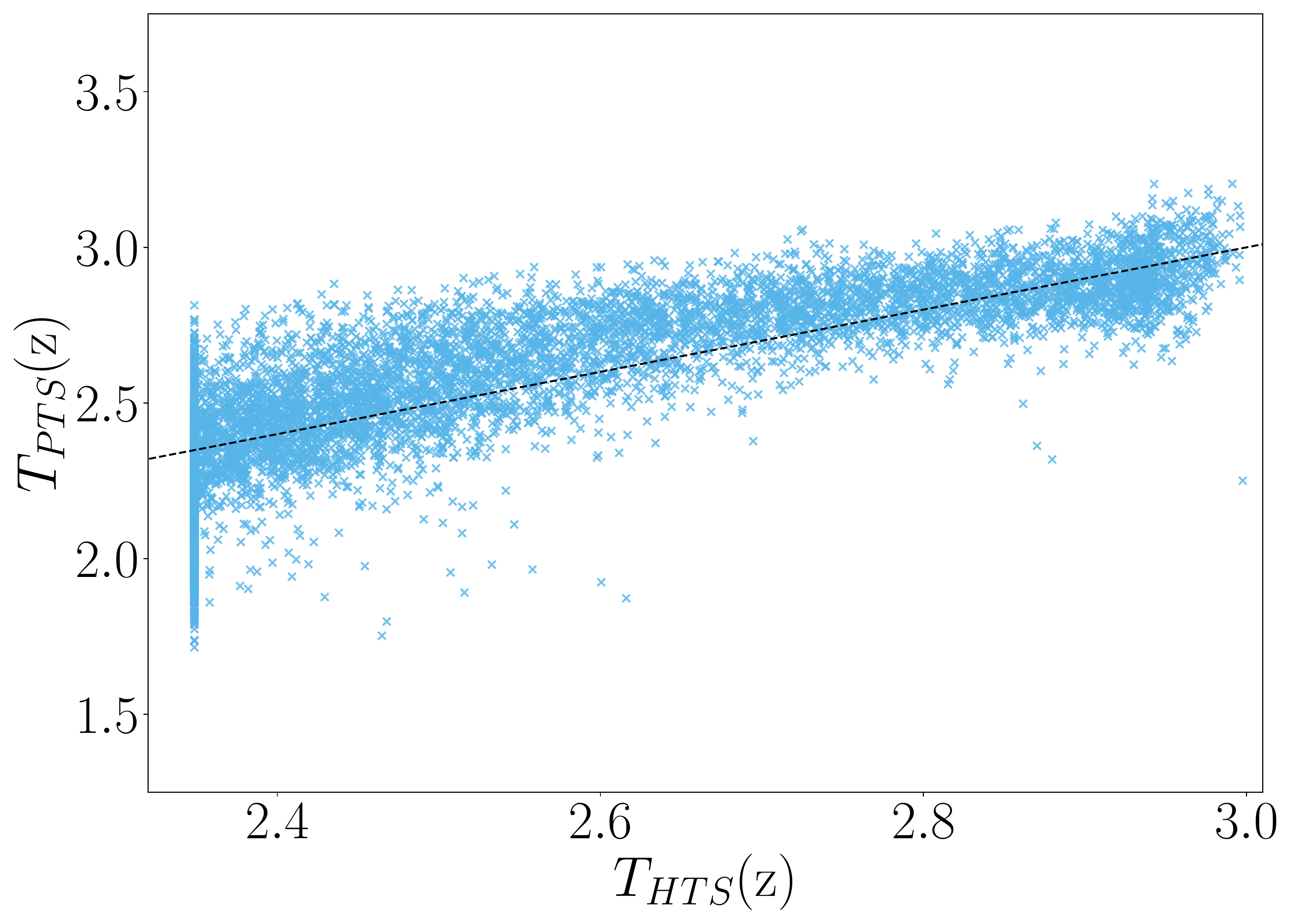}
     \caption{Temperature factor computed by PTS against temperature factor computed by HTS for test set predictions. The black dotted line shows represents the one to one relation.}
     \label{fig:PTSvHTS}
 \end{figure}

In Figure \ref{fig:h_PTS} we show that a much more expressive method like PTS also captures this linear relationship when given enough data. However, in the face of limited data, it fails to do so. Moreover, in Figure \ref{fig:PTSvHTS} we plot for all samples in the test set the temperature factors given by PTS against those by HTS, both methods fitted using all validation samples. The plot shows that when data is plentiful the function learnt by both methods is reasonable similar, suggesting that the function space of HTS contains well-performing solutions similar to those learnt by PTS despite being much more constrained.

\begin{figure}
    \centering
    \begin{subfigure}{.475\textwidth}
         \centering
         \includegraphics[width=\linewidth]{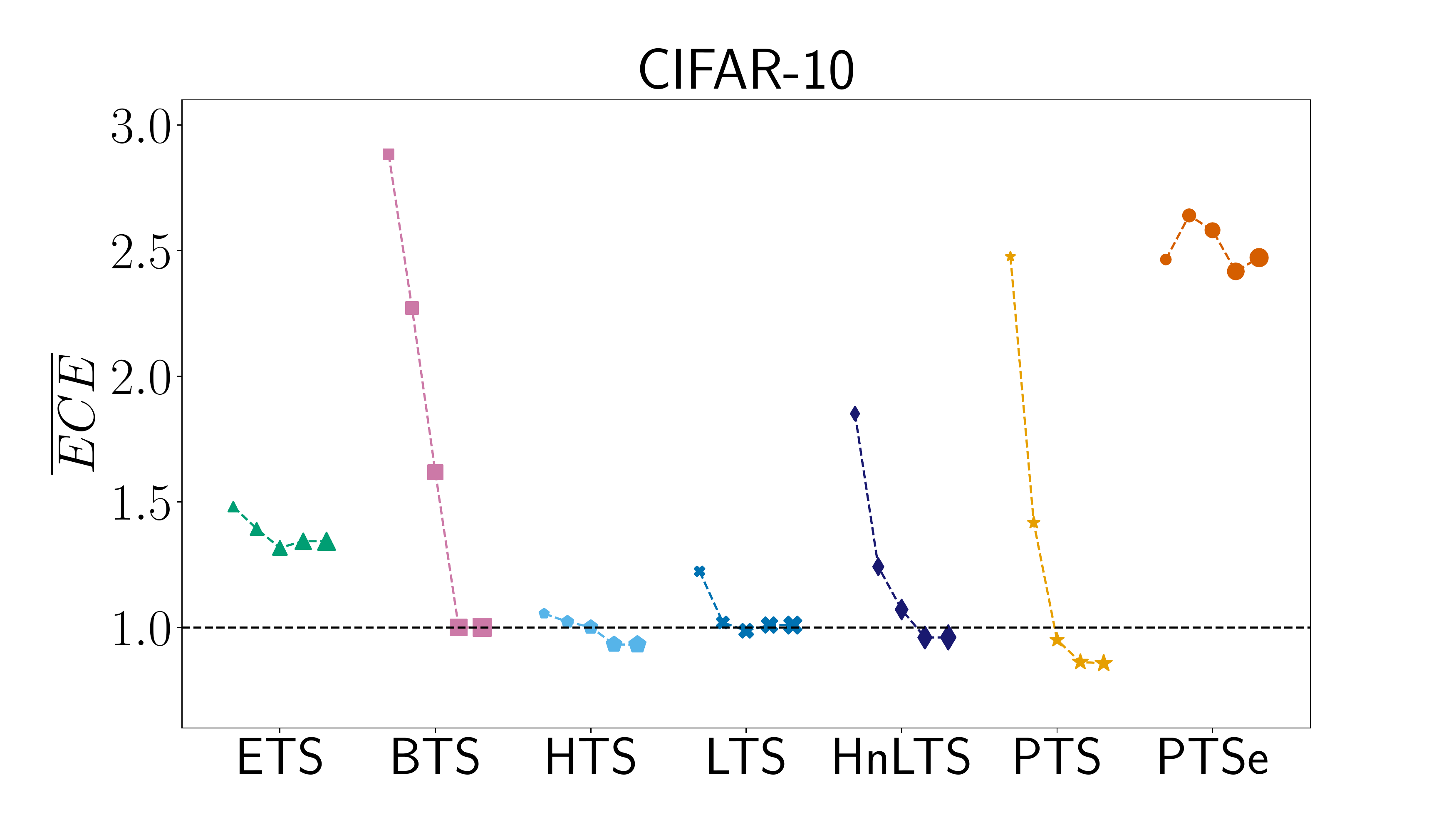}
         \caption{Average Relative ECE in CIFAR-10 tasks.}
         \label{fig:relECEc10}
     \end{subfigure}
     \begin{subfigure}{.475\textwidth}
         \centering
         \includegraphics[width=\linewidth]{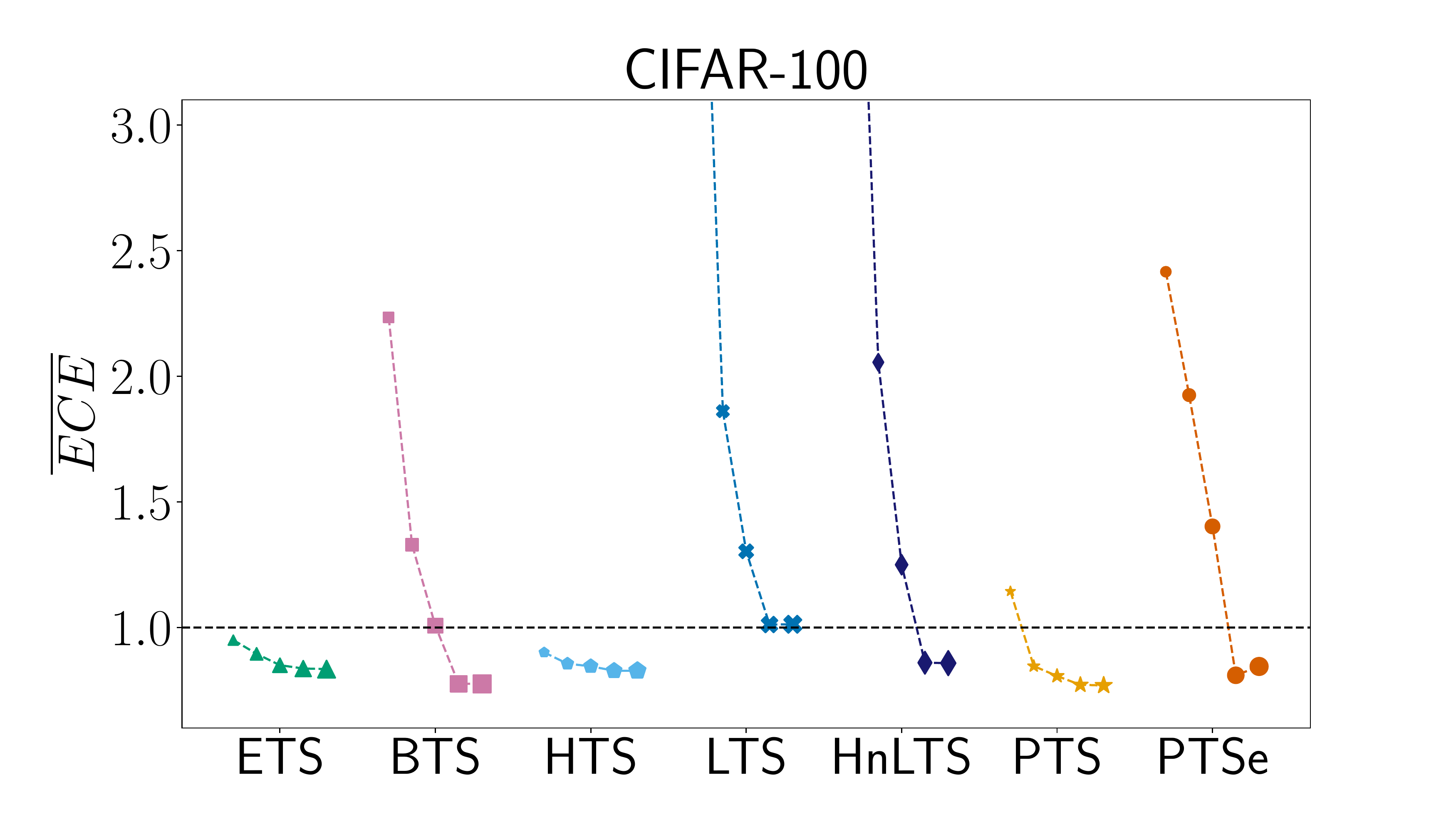}
         \caption{Average Relative ECE in CIFAR-100 tasks.}
         \label{fig:relECEc100}
     \end{subfigure}

     \begin{subfigure}{.475\textwidth}
         \centering
         \includegraphics[width=\linewidth]{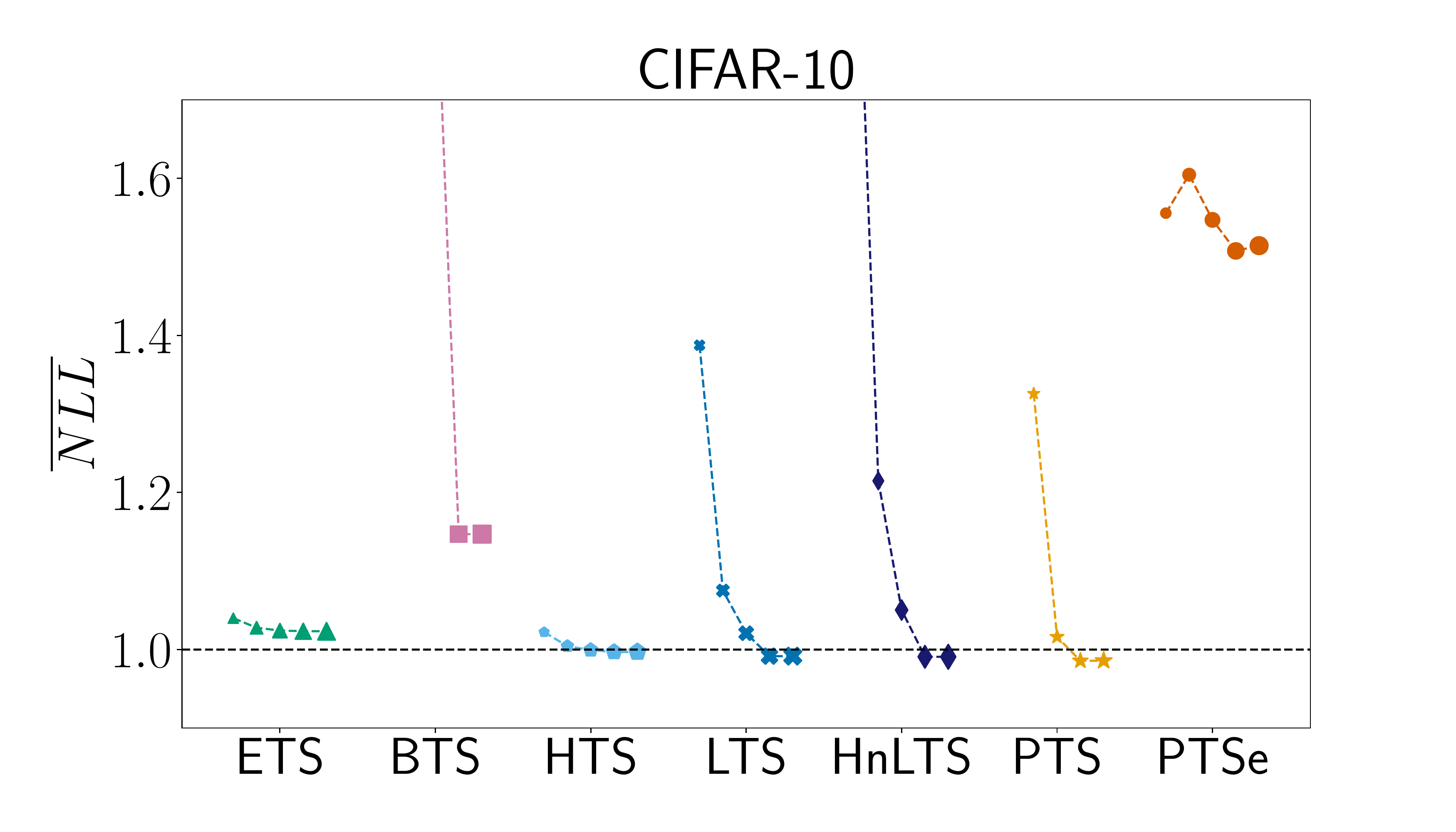}
         \caption{Average Relative NLL in CIFAR-10 tasks.}
         \label{fig:relNLLc10}
     \end{subfigure}
     \begin{subfigure}{.475\textwidth}
         \centering
         \includegraphics[width=\linewidth]{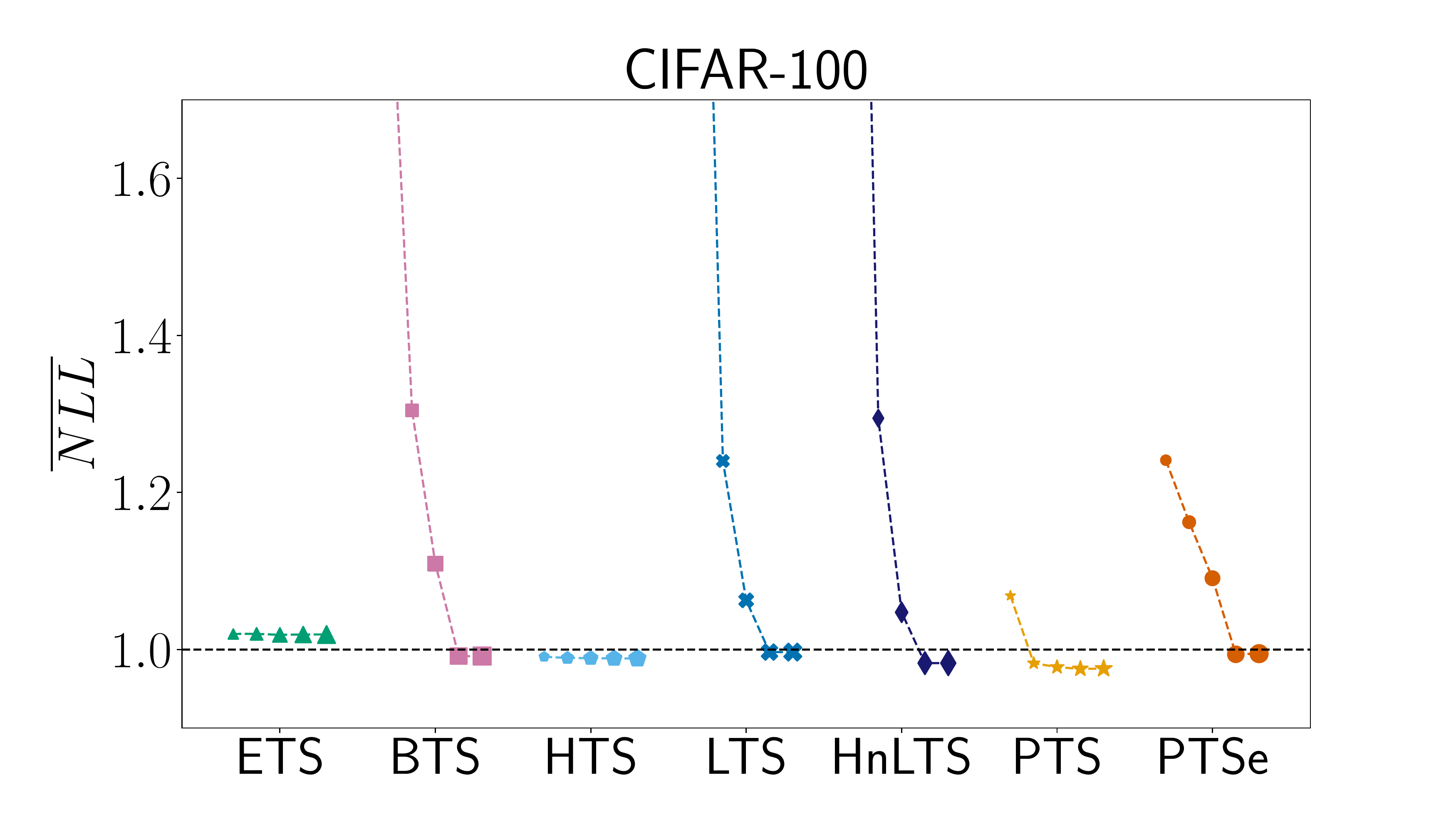}
         \caption{Average Relative NLL in CIFAR-100 tasks.}
         \label{fig:relNLLc100}
     \end{subfigure}
    
    \caption{Average results for CIFAR-10 (left) and CIFAR-100 (right) tasks of all calibration methods in terms of ECE (up) and NLL (down) normalized by the performance of TS, namely $\overline{ECE}$ and $\overline{NLL}$.}
         \label{fig:avgres}

\end{figure}

\subsection{Benchmarking}\label{subsec:bench}

In this section, we compare the performance of the proposed ATS methods: LTS, HTS, and HnLTs; with state-of-the-art accuracy-preserving methods: TS, ETS, BTS, and PTS. We fit two versions of PTS: One trained to minimize the NLL, the calibration objective we use to train every method; and a second version optimizing the ECE-based loss instead as reported in \cite{PTS}(see Section \ref{sec:base}). We refer to the former as PTS and the latter as PTSe where the `e' stands for the ECE-based objective.

\subsubsection{Results}

For the sake of space and simplicity, we depict results for each dataset and average across models---e.g. average ECE of HTS on all CIFAR-10 tasks---. We defer detailed results to \ref{apx:res}. Results are shown in Figure \ref{fig:avgres}. We normalize each metric by the performance of TS as we consider it the main benchmark. We report performance in terms of normalized ECE and normalized NLL, namely $\overline{ECE}$ and $\overline{NLL}$. For each method, we plot five markers, the size of which increases with the size of the validation data set. From smallest to biggest these are $N = (200, 500, 1000, 5000, 10000)$. The $y$-axis position of the marker indicates the mean value across tasks, where each task is a different NN architecture calibrated.

We first point out that almost all models outperform the simple TS when there is enough data (big markers), although, on average, there are no big differences between models. However, when data is scarce all the highly-parametrized models show severe performance degradation and only ETS and HTS seem to provide consistent performance. Moreover HTS provides better results in most of the individual tasks while ETS barely outperforms the baseline TS.

Also, it is worth noting the difference between datasets. In the highly dimensional CIFAR100, we can see a greater advantage in using calibration methods more complex than TS. On the other hand, the best methods barely outperform TS in CIFAR10 tasks. This suggests that the problem of calibration may grow more complex with the number of classes, although the number of datasets included in our experiments is not representative enough and more experiments are required to validate this observation.

Interestingly, HnLTS fails in low-data scenarios, even though it could, in theory, recover the HTS solution by zeroing the $w^L$ parameter. This suggests that increasing expressiveness can do more harm than good by complicating the training objective.


\section{Conclusions}

We have shown that post-hoc calibration of DNNs can benefit from more expressive models than the widely used Temperature Scaling, especially in tasks with a high number of classes. For instance, simply adjusting the temperature factor of TS with a linear combination of the logit prediction improves calibration by taking into account the score assigned to each class.

However, more complex models require higher amounts of data to find a good-performing solution. This poses a trade-off between the complexity of the calibration model and the available data to train the model. There are many real-world tasks where data for re-calibration is limited and hinders the calibration with a complex model. 

By analysing the calibration functions learned by expressive models on plenty of data, we can design simpler models with a strong inductive bias towards similar calibration functions. In this work, we have introduced HTS, a 2-parameter model that scales predictions according to their entropy. The temperature factors estimated by PTS, a much more expressive model, follow the same linear relation with the predictive entropy that HTS implicitly assumes. HTS shows calibration performance comparable to that of more expressive methods on ideal data conditions. However, unlike other methods, it is robust to data scarcity. Moreover, an important feature of the model is that it is interpretable, characterizing the link between a prediction's uncertainty and its over-confidence.

With this work, we motivate the study of expressive methods as a way to design practical models with a suitable inductive bias. As a first approach, we propose to use a hand-designed low-parameter model to achieve this bias. In future work, we plan to try other forms of inducing the desired bias, for instance, via the prior specification in a Bayesian inference setting. This option may allow training higher capacity models while still being robust to data scarcity.

\section*{Acknowledgements} 
    D. R. and S. A. B. are funded by the Spanish Ministerio de Ciencia, Innovación y Universidades (MCIU), the Spanish Agencia Estatal de Investigación (AEI) and the Fondo Europeo de Desarrollo Regional (FEDER, UE); under project RTI2018-098091-B-I00.

    Juan Maroñas acknowledges funding coming from the Spanish National Research Project PID2019-106827GB-I00.

\clearpage
\appendix

\section{Results}\label{apx:res}

In this section, we provide in tables the results for each model-dataset task. Additionally, we give average performance normalized by that of the uncalibrated model across tasks in each dataset.

Results of ECE ($M=50$), NLL, and Brier score, using the whole validation set are shown in Table \ref{tab:ece}, Table \ref{tab:nll}, and Table \ref{tab:bri}, respectively. Table \ref{tab:ece_mhd}, Table \ref{tab:nll_mhd}, and Table \ref{tab:bri_mhd}, show average results using $5000$ validation samples, randomly chosen at each experiment run, to calibrate models. Equivalently, tables \ref{tab:ece_md}, \ref{tab:nll_md}, and \ref{tab:bri_md}, show the same results but using $1000$ validation samples; tables \ref{tab:ece_mld}, \ref{tab:nll_mld}, and \ref{tab:bri_mld} show average results using $500$ validation samples. Lastly, tables \ref{tab:ece_ld}, \ref{tab:nll_ld}, and \ref{tab:bri_ld} show average results using $200$ validation samples.

\begin{table}[hb]
    \centering
    \caption{ECE ($M=50$). Models are denoted by their architecture and depth (and width if applicable).}
    \label{tab:ece}
    \resizebox{\textwidth}{!}{\begin{tabular}{llccccccccc}
    \toprule
    &  Task  &  Uncalibrated &   TS &  ETS  &  BTS &  HTS & LTS &  HnLTS & PTS & \begin{tabular}{@{}c@{}}PTS \\ ($L_{ECE}$)\end{tabular}  \\
    \midrule
    \multirow{10}{*}{\rotatebox[origin=c]{90}{CIFAR 10}} 
    & DenseNet 121      &          2.84 & 1.39 & 2.26 & \textbf{1.09} &       1.41 &       1.30 &         1.35 & 1.13 &     3.20 \\
    & DenseNet 169      &          2.79 & 1.65 & 2.31 & 1.27 &       1.10 &       1.29 &         1.02 & \textbf{0.91} &     2.18 \\
    & ResNet 50         &         10.71 & 2.54 & 2.84 & 1.86 &       \textbf{1.62} &       2.43 &         2.12 & 1.66 &     6.61 \\
    & ResNet 101        &          4.45 & \textbf{1.37} & 1.49 & 1.72 &       1.54 &       1.42 &         1.47 & 1.43 &     4.20 \\
    & ResNext 29 8x16   &          2.88 & 1.03 & \textbf{0.85} & 1.19 &       1.06 &       1.11 &         1.15 & 1.01 &     3.39 \\
    & VGG 19            &          4.61 & 2.51 & 3.30 & \textbf{1.38} &       1.62 &       2.15 &         1.67 & 1.49 &     5.02 \\
    & WRN 28x10         &          1.93 & \textbf{0.70} & 1.59 & 1.07 &       0.88 &       0.88 &         0.87 & 0.86 &     1.25 \\
    & WRN 40x10         &          3.12 & 1.04 & 1.14 & 1.28 &       1.11 &       1.21 &         1.21 & \textbf{1.03} &     3.55 \\
    \cline{2-11}
    \\[-1em]
    & Avg. Relative $\overline{ECE}$ &          1.00 & 0.40 & 0.55 & 0.38 & 0.37 & 0.40 & 0.37 & \textbf{0.33} & 0.94 \\
    \midrule
    \multirow{ 9}{*}{\rotatebox[origin=c]{90}{CIFAR 100}} 
    & DenseNet 121      &          8.76 & 3.93 & 2.96 & \textbf{2.77} &       3.27 &       4.12 &         3.23 & 3.22 &     3.26 \\
    & DenseNet 169      &          8.93 & 3.95 & 2.85 & 3.19 &       3.36 &       4.15 &         3.67 & 3.21 &     \textbf{2.76} \\
    & ResNet 101        &         11.45 & 2.25 & 2.22 & 2.22 &       2.32 &       2.66 &         2.32 & \textbf{1.92} &     2.44 \\
    & ResNext 29 8x16   &          9.69 & 3.14 & 2.80 & 2.06 &       2.09 &       3.51 &         2.55 & \textbf{1.79} &     2.79 \\
    & VGG 19            &         17.63 & 5.13 & 5.36 & 3.89 &       3.78 &       3.60 &         3.56 & \textbf{3.28} &     4.22 \\
    & WRN 28x10         &          5.19 & 4.63 & 3.64 & \textbf{3.11} &       3.52 &       4.59 &         3.69 & 3.33 &     4.10 \\
    & WRN 40x10         &         14.78 & 4.20 & \textbf{2.76} & 3.55 &       3.85 &       4.19 &    3.87 & 4.09 &     2.95 \\
    \cline{2-11}
    \\[-1em]
    & Avg. Relative $\overline{ECE}$ &          1.00 & 0.41 & 0.33 & \textbf{0.31} & 0.33 & 0.41 & 0.35 & 0.31 & 0.34 \\
    \bottomrule
\end{tabular}}
\end{table}

\begin{table}
    \centering
    \caption{NLL. Models are denoted by their architecture and depth (and width if applicable).}
    \label{tab:nll}
    \resizebox{\textwidth}{!}{\begin{tabular}{llccccccccc}
    \toprule
    &  Task  &  Uncalibrated &   TS &  ETS  &  BTS &  HTS & LTS &  HnLTS & PTS
    & \begin{tabular}{@{}c@{}}PTS \\ ($L_{ECE}$)\end{tabular}  \\
    \midrule
    \multirow{ 10}{*}{\rotatebox[origin=c]{90}{CIFAR 10}} 
    & DenseNet 121      &          0.1881 & 0.1618 & 0.1673  & 0.1835 &     0.1611 &     0.1606 &       0.1604 & \textbf{0.1585} &   0.2407 \\
    & DenseNet 169      &          0.1870 & 0.1608 & 0.1686  & 0.2035 &     0.1590 &     0.1593 &       0.1584 & \textbf{0.1542} &   0.1934 \\
    & ResNet 50         &          0.7897 & 0.4473 & 0.4515  & 0.4498 &     0.4447 &     0.4399 &       \textbf{0.4385} & 0.4444 &   0.6847 \\
    & ResNet 101        &          0.3047 & 0.2163 & 0.2199  & 0.2411 &     0.2173 &     \textbf{0.2132} &       0.2142 & 0.2168 &   0.3570 \\
    & ResNext 29 8x16   &          0.1997 & 0.1621 & \textbf{0.1617}  & 0.1909 &     0.1623 &     0.1635 &       0.1634 & 0.1628 &   0.2820 \\
    & VGG 19            &          0.2998 & 0.2355 & 0.2424  & 0.2672 &     0.2330 &     0.2313 &       0.2299 & 0.\textbf{2296} &   0.4191 \\
    & WRN 28x10         &          0.1497 & 0.1362 & 0.1424  & 0.1750 &     0.1364 &     0.1348 &       0.1355 & \textbf{0.1346} &   0.1390 \\
    & WRN 40x10         &          0.2068 & 0.1634 & 0.1642  & 0.1729 &     0.1634 &     0.1632 &       0.1633 & \textbf{0.1607} &   0.2783 \\
    \cline{2-11}
    \\[-1em]
    & Avg. Relative $\overline{NLL}$ &          1.0000 & 0.7876 & 0.8060  & 0.9096 & 0.7842 & 0.7804 & 0.7798 & \textbf{0.7750} & 1.1797 \\
    \midrule
    \multirow{ 9}{*}{\rotatebox[origin=c]{90}{CIFAR 10}} 
    & DenseNet 121      &          0.8939 & 0.8355 & 0.8589  & 0.8271 &     0.8253 &     0.8351 &       0.8209 & \textbf{0.8161} &   0.8321 \\
    & DenseNet 169      &          0.8748 & 0.8156 & 0.8328  & 0.8082 &     0.8061 &     0.8152 &       0.8021 & \textbf{0.7950} &   0.8112 \\
    & ResNet 101        &          1.1343 & 1.0007 & 1.0088  & 1.0040 &     1.0002 &     1.0061 &       1.0035 & \textbf{0.9971} &   1.0037 \\
    & ResNext 29 8x16   &          0.9398 & 0.8220 & 0.8404  & 0.8190 &     0.8128 &     0.8282 &       0.8182 & \textbf{0.8044} &   0.8173 \\
    & VGG 19            &          1.5414 & 1.1997 & 1.2069  & 1.1926 &     1.1941 &     1.1825 &       1.1821 & \textbf{1.1737} &   1.1939 \\
    & WRN 28x10         &          0.8173 & 0.8135 & 0.8343  & 0.7876 &     0.7868 &     0.8040 &       0.7765 & \textbf{0.7719} &   0.8054 \\
    & WRN 40x10         &          1.2248 & 0.9055 & 0.9261  & 0.9037 &     0.8987 &     0.8969 &       \textbf{0.8821} & 0.8832 &   0.8959 \\
    \cline{2-11}
    \\[-1em]
    & Avg. Relative $\overline{NLL}$ &          1.0000 & 0.8767 & 0.8938  & 0.8687 & 0.8661 & 0.8739 & 0.8611 & \textbf{0.8548} & 0.8720 \\
    \bottomrule
\end{tabular}}
\end{table}

\begin{table}
    \centering
    \caption{Brier Score. Models are denoted by their architecture and depth (and width if applicable).}
    \label{tab:bri}
    \resizebox{\textwidth}{!}{\begin{tabular}{llccccccccc}
    \toprule
    &  Task  &  Uncalibrated &   TS &  ETS &  BTS &  HTS & LTS &  HnLTS & PTS & \begin{tabular}{@{}c@{}}PTS \\ ($L_{ECE}$)\end{tabular}  \\
    \midrule
    \multirow{ 10}{*}{\rotatebox[origin=c]{90}{CIFAR 10}} 
    & DenseNet 121      &          0.0764 & 0.0729 & 0.0724 & 0.0722 &     0.0726 &     0.0727 &       0.0724 & \textbf{0.0720} &   0.0787 \\
    & DenseNet 169      &          0.0754 & 0.0717 & 0.0712 & 0.0711 &     0.0712 &     0.0714 &       0.0710 & \textbf{0.0705} &   0.0740 \\
    & ResNet 50         &          0.2392 & 0.2037 & 0.2029 & 0.2036 &     0.2027 &     0.2020 &       \textbf{0.2014} & 0.2025 &   0.2233 \\
    & ResNet 101        &          0.1102 & 0.1011 & 0.1010 & 0.1019 &     0.1011 &     0.1004 &       \textbf{0.1004} & 0.1007 &   0.1113 \\
    & ResNext 29 8x16   &          0.0828 & 0.0783 & 0.0782 & 0.0791 &     0.0784 &     0.0783 &       0.0783 & \textbf{0.0781} &   0.0869 \\
    & VGG 19            &          0.1101 & 0.1019 & 0.1007 & 0.1003 &     0.1005 &     0.1011 &       0.1000 & \textbf{0.0999} &   0.1139 \\
    & WRN 28x10         &          0.0629 & 0.0608 & 0.0609 & 0.0611 &     0.0607 &     0.0606 &       \textbf{0.0606} & 0.0606 &   0.0612 \\
    & WRN 40x10         &          0.0820 & 0.0768 & 0.0764 & 0.0770 &     0.0767 &     0.0767 &       0.0767 & \textbf{0.0764} &   0.0853 \\
    \cline{2-11}
    \\[-1em]
    & Avg. Relative $\overline{Brier}$ &          1.0000 & 0.9312 & 0.9275 & 0.9304 & 0.9275 & 0.9272 & 0.9245 & \textbf{0.9231} & 1.0069 \\
    \midrule
    \multirow{ 9}{*}{\rotatebox[origin=c]{90}{CIFAR 100}} 
    & DenseNet 121      &          0.3171 & 0.3048 & 0.3043 & 0.3052 &     0.3049 &     0.3040 &       0.3036 & \textbf{0.3024} &   0.3029 \\
    & DenseNet 169      &          0.3142 & 0.3017 & 0.3010 & 0.3020 &     0.3016 &     0.2999 &       0.2995 & \textbf{0.2984} &   0.2990 \\
    & ResNet 101        &          0.4053 & 0.3817 & 0.3816 & 0.3825 &     0.3817 &     0.3817 &       0.3814 & \textbf{0.3805} &   0.3817 \\
    & ResNext 29 8x16   &          0.3275 & 0.3096 & 0.3093 & 0.3096 &     0.3090 &     0.3100 &       0.3097 & \textbf{0.3068} &   0.3085 \\
    & VGG 19            &          0.4433 & 0.3918 & 0.3910 & 0.3880 &     0.3897 &     0.3882 &       0.3878 & \textbf{0.3848} &   0.3868 \\
    & WRN 28x10         &          0.2892 & 0.2886 & 0.2877 & 0.2871 &     0.2872 &     0.2851 &       0.2839 & \textbf{0.2831} &   0.2868 \\
    & WRN 40x10         &          0.3700 & 0.3276 & 0.3270 & 0.3293 &     0.3288 &     0.3240 &       0.3240 & 0.3245 &   \textbf{0.3239} \\
    \cline{2-11}
    \\[-1em]
    & Avg. Relative $\overline{Brier}$ &          1.0000 & 0.9394 & 0.9378 & 0.9387 & 0.9382 & 0.9341 & 0.9328 & \textbf{0.9289} & 0.9328 \\
    \bottomrule
\end{tabular}}
\end{table}


\begin{table}
    \centering
    \caption{ECE ($M=50$) using $5000$ validation samples. Models are denoted by their architecture and depth (and width if applicable).}
    \label{tab:ece_mhd}
    \resizebox{\textwidth}{!}{\begin{tabular}{llccccccccc}
    \toprule
    & Task  &  Uncalibrated &   TS &  ETS &  BTS &  HTS & LTS &  HnLTS & PTS & \begin{tabular}{@{}c@{}}PTS \\ ($L_{ECE}$)\end{tabular}  \\
    \midrule
    \multirow{ 10}{*}{\rotatebox[origin=c]{90}{CIFAR 10}} 
    & DenseNet 121      &          2.84 & 1.39 & 2.26 & \textbf{1.09} &       1.41 &       1.29 &         1.34 & 1.10 &     2.99 \\
    & DenseNet 169      &          2.79 & 1.65 & 2.31 & 1.27 &       1.10 &       1.29 &         1.02 & \textbf{0.97} &     2.33 \\
    & ResNet 50         &         10.71 & 2.54 & 2.84 & 1.86 &       \textbf{1.62} &       2.43 &         2.14 & 1.70 &     5.53 \\
    & ResNet 101        &          4.45 & \textbf{1.37} & 1.49 & 1.72 &       1.54 &       1.42 &         1.47 & 1.43 &     4.46 \\
    & ResNext 29 8x16   &          2.88 & 1.03 & \textbf{0.85} & 1.19 &       1.06 &       1.11 &         1.14 & 1.03 &     3.12 \\
    & VGG 19            &          4.61 & 2.51 & 3.30 & \textbf{1.38} &       1.62 &       2.15 &         1.67 & 1.41 &     5.08 \\
    & WRN 28x10         &          1.93 & \textbf{0.70} & 1.59 & 1.07 &       0.89 &       0.89 &         0.86 & 0.88 &     1.24 \\
    & WRN 40x10         &          3.12 & 1.04 & 1.14 & 1.28 &       1.11 &       1.21 &         1.22 & \textbf{1.02} &     3.64 \\
    \cline{2-11}
    \\[-1em]
    & Avg. Relative $\overline{ECE}$ &          1.00 & 0.40 & 0.55 & 0.38 & 0.37 & 0.40 & 0.37 & \textbf{0.33} & 0.93 \\
    \midrule
    \multirow{ 9}{*}{\rotatebox[origin=c]{90}{CIFAR 100}} 
    & DenseNet 121      &          8.76 & 3.93 & 2.95 & \textbf{2.77} &       3.28 &       4.11 &         3.23 & 3.19 &     3.10 \\
    & DenseNet 169      &          8.93 & 3.95 & \textbf{2.85} & 3.19 &       3.36 &       4.16 &         3.67 & 3.20 &     2.91 \\
    & ResNet 101        &         11.45 & 2.25 & 2.23 & 2.22 &       2.30 &       2.65 &         2.34 & \textbf{1.90} &     2.41 \\
    & ResNext 29 8x16   &          9.69 & 3.14 & 2.83 & 2.06 &       2.09 &       3.50 &         2.56 & \textbf{2.01} &     2.45 \\
    & VGG 19            &         17.63 & 5.13 & 5.36 & 3.89 &       3.79 &       3.58 &         3.58 & \textbf{3.28} &     3.82 \\
    & WRN 28x10         &          5.19 & 4.63 & 3.65 & \textbf{3.11} &       3.54 &       4.60 &         3.68 & 3.23 &     4.08 \\
    & WRN 40x10         &         14.78 & 4.20 & \textbf{2.77} & 3.55 &       3.86 &       4.19 &         3.87 & 4.03 &     2.81 \\
    \cline{2-11}
    \\[-1em]
    & Avg. Relative $\overline{ECE}$ &          1.00 & 0.41 & 0.33 & \textbf{0.30} & 0.33 & 0.41 & 0.35 & 0.31 & 0.33 \\
    \bottomrule
\end{tabular}}
\end{table}

\begin{table}
    \centering
    \caption{NLL using $5000$ validation samples. Models are denoted by their architecture and depth (and width if applicable).}
    \label{tab:nll_mhd}
    \resizebox{\textwidth}{!}{\begin{tabular}{llccccccccc}
    \toprule
    &  Task  &  Uncalibrated &   TS &  ETS  &  BTS &  HTS & LTS &  HnLTS & PTS
    & \begin{tabular}{@{}c@{}}PTS \\ ($L_{ECE}$)\end{tabular}  \\
    \midrule
    \multirow{ 10}{*}{\rotatebox[origin=c]{90}{CIFAR 10}} 
    & DenseNet 121      &          0.1881 & 0.1618 & 0.1673 & 0.1835 &     0.1611 &     0.1606 &       0.1604 & \textbf{0.1582} &   0.2267 \\
    & DenseNet 169      &          0.1870 & 0.1608 & 0.1686 & 0.2035 &     0.1590 &     0.1593 &       0.1584 & \textbf{0.1545} &   0.2111 \\
    & ResNet 50         &          0.7897 & 0.4473 & 0.4515 & 0.4498 &     0.4447 &     0.4399 &       \textbf{0.4385} & 0.4446 &   0.7042 \\
    & ResNet 101        &          0.3047 & 0.2163 & 0.2199 & 0.2411 &     0.2173 &     \textbf{0.2132} &       0.2142 & 0.2169 &   0.3689 \\
    & ResNext 29 8x16   &          0.1997 & 0.1621 & \textbf{0.1617} & 0.1909 &     0.1623 &     0.1635 &       0.1634 & 0.1627 &   0.2557 \\
    & VGG 19            &          0.2998 & 0.2355 & 0.2424 & 0.2672 &     0.2330 &     0.2313 &       0.2299 & \textbf{0.2291} &   0.4007 \\
    & WRN 28x10         &          0.1497 & 0.1362 & 0.1424 & 0.1750 &     0.1364 &     0.1348 &       0.1355 & \textbf{0.1347} &   0.1423 \\
    & WRN 40x10         &          0.2068 & 0.1634 & 0.1642 & 0.1729 &     0.1634 &     0.1632 &       0.1633 & \textbf{0.1608} &   0.2849 \\
    \cline{2-11}
    \\[-1em]
    & Avg. Relative $\overline{NLL}$ &          1.0000 & 0.7867 & 0.8060 & 0.9096 & 0.7842 & 0.7804 & 0.7798 & \textbf{0.7748} & 1.1727 \\
    \midrule
    \multirow{ 9}{*}{\rotatebox[origin=c]{90}{CIFAR 100}} 
    & DenseNet 121      &          0.8939 & 0.8355 & 0.8587 & 0.8271 &     0.8253 &     0.8351 &       0.8209 & \textbf{0.8152} &   0.8331 \\
    & DenseNet 169      &          0.8748 & 0.8156 & 0.8328 & 0.8082 &     0.8061 &     0.8152 &       0.8021 & \textbf{0.7943} &   0.8104 \\
    & ResNet 101        &          1.1343 & 1.0007 & 1.0088 & 1.0040 &     1.0002 &     1.0060 &       1.0035 & \textbf{0.9972} &   1.0044 \\
    & ResNext 29 8x16   &          0.9398 & 0.8220 & 0.8404 & 0.8190 &     0.8128 &     0.8282 &       0.8182 & \textbf{0.8065} &   0.8144 \\
    & VGG 19            &          1.5414 & 1.1997 & 1.2069 & 1.1926 &     1.1941 &     1.1826 &       1.1821 & \textbf{1.1751} &   1.1903 \\
    & WRN 28x10         &          0.8173 & 0.8135 & 0.8342 & 0.7876 &     0.7868 &     0.8041 &       0.7765 & \textbf{0.7691} &   0.8069 \\
    & WRN 40x10         &          1.2248 & 0.9055 & 0.9261 & 0.9037 &     0.8987 &     0.8969 &       \textbf{0.8821} & 0.8828 &   0.8943 \\
    \cline{2-11}
    \\[-1em]
    & Avg. Relative $\overline{NLL}$ &          1.0000 & 0.8767 & 0.8937 & 0.8687 & 0.8661 & 0.8739 & 0.8611 & \textbf{0.8545} & 0.8714 \\
    \bottomrule
\end{tabular}}
\end{table}

\begin{table}
    \centering
    \caption{Brier Score using $5000$ validation samples. Models are denoted by their architecture and depth (and width if applicable).}
    \label{tab:bri_mhd}
    \resizebox{\textwidth}{!}{\begin{tabular}{llccccccccc}
    \toprule
    &  Task  &  Uncalibrated &   TS &  ETS &  BTS &  HTS & LTS &  HnLTS & PTS & \begin{tabular}{@{}c@{}}PTS \\ ($L_{ECE}$)\end{tabular}  \\
    \midrule
    \multirow{10}{*}{\rotatebox[origin=c]{90}{CIFAR 10}} 
    & DenseNet 121      &          0.0764 & 0.0729 & 0.0724 & 0.0722 &     0.0726 &     0.0727 &       0.0724 & \textbf{0.0720} &   0.0778 \\
    & DenseNet 169      &          0.0754 & 0.0717 & 0.0712 & 0.0711 &     0.0712 &     0.0714 &       0.0710 & \textbf{0.0706} &   0.0747 \\
    & ResNet 50         &          0.2392 & 0.2037 & 0.2029 & 0.2036 &     0.2026 &     0.2020 &       \textbf{0.2014} & 0.2026 &   0.2199 \\
    & ResNet 101        &          0.1102 & 0.1011 & 0.1010 & 0.1019 &     0.1011 &     0.1004 &       \textbf{0.1004} & 0.1007 &   0.1123 \\
    & ResNext 29 8x16   &          0.0828 & 0.0783 & 0.0782 & 0.0791 &     0.0784 &     0.0783 &       0.0783 & \textbf{0.0781} &   0.0856 \\
    & VGG 19            &          0.1101 & 0.1019 & 0.1007 & 0.1003 &     0.1005 &     0.1011 &       0.1000 & \textbf{0.0998} &   0.1139 \\
    & WRN 28x10         &          0.0629 & 0.0608 & 0.0609 & 0.0611 &     0.0607 &     0.0606 &       \textbf{0.0606} & 0.0606 &   0.0614 \\
    & WRN 40x10         &          0.0820 & 0.0768 & 0.0764 & 0.0770 &     0.0767 &     0.0767 &       0.0767 & \textbf{0.0764} &   0.0858 \\
    \cline{2-11}
    \\[-1em]
    & Avg. Relative $\overline{Brier}$ &          1.0000 & 0.9312 & 0.9275 & 0.9304 & 0.9275 & 0.9272 & 0.9245 & \textbf{0.9231} & 1.0048 \\
    \midrule
    \multirow{ 9}{*}{\rotatebox[origin=c]{90}{CIFAR 100}}  
    & DenseNet 121      &          0.3171 & 0.3048 & 0.3043 & 0.3052 &     0.3049 &     0.3040 &       0.3036 & \textbf{0.3023} &   0.3029 \\
    & DenseNet 169      &          0.3142 & 0.3017 & 0.3010 & 0.3020 &     0.3016 &     0.2999 &       0.2995 & \textbf{0.2982} &   0.2991 \\
    & ResNet 101        &          0.4053 & 0.3817 & 0.3816 & 0.3825 &     0.3817 &     0.3817 &       0.3814 & \textbf{0.3805} &   0.3818 \\
    & ResNext 29 8x16   &          0.3275 & 0.3096 & 0.3093 & 0.3096 &     0.3090 &     0.3100 &       0.3097 & \textbf{0.3071} &   0.3079 \\
    & VGG 19            &          0.4433 & 0.3918 & 0.3910 & 0.3880 &     0.3897 &     0.3882 &       0.3878 & 0.3851 &   \textbf{0.3850} \\
    & WRN 28x10         &          0.2892 & 0.2886 & 0.2877 & 0.2871 &     0.2872 &     0.2851 &       0.2839 & \textbf{0.2827} &   0.2871 \\
    & WRN 40x10         &          0.3700 & 0.3276 & 0.3270 & 0.3293 &     0.3288 &     0.3240 &       0.3240 & 0.3245 &   \textbf{0.3232} \\
    \cline{2-11}
    \\[-1em]
    & Avg. Relative $\overline{Brier}$ &          1.0000 & 0.9394 & 0.9378 & 0.9387 & 0.9382 & 0.9341 & 0.9328 & \textbf{0.9289} &  0.9320 \\
    \bottomrule
\end{tabular}}
\end{table}

\begin{table}
    \centering
    \caption{ECE ($M=50$) using $1000$ validation samples. Models are denoted by their architecture and depth (and width if applicable).}
    \label{tab:ece_md}
    \resizebox{\textwidth}{!}{\begin{tabular}{llccccccccc}
    \toprule
    & Task  &  Uncalibrated &   TS &  ETS &  BTS &  HTS & LTS &  HnLTS & PTS & \begin{tabular}{@{}c@{}}PTS \\ ($L_{ECE}$)\end{tabular}  \\
    \midrule
    \multirow{ 10}{*}{\rotatebox[origin=c]{90}{CIFAR 10}} 
    & DenseNet 121      &          2.84 & 1.52 & 2.29 & 2.12 &       1.51 &       \textbf{1.30} &         1.49 & 1.34 &     3.09 \\
    & DenseNet 169      &          2.79 & 1.55 & 2.40 & 1.79 &       1.32 &       \textbf{1.20} &         1.35 & 1.22 &     3.08 \\
    & ResNet 50         &         10.71 & 2.56 & 2.75 & 3.19 &       2.16 &       2.38 &         2.22 & \textbf{1.90} &     7.79 \\
    & ResNet 101        &          4.45 & \textbf{1.46} & 1.57 & 2.72 &       1.66 &       1.64 &         1.73 & 1.53 &     4.09 \\
    & ResNext 29 8x16   &          2.88 & \textbf{1.04} & 1.08 & 2.24 &       1.28 &       1.29 &         1.40 & 1.25 &     3.16 \\
    & VGG 19            &          4.61 & 2.54 & 3.20 & 2.65 &       1.99 &       1.99 &         1.95 & \textbf{1.72} &     4.88 \\
    & WRN 28x10         &          1.93 & \textbf{0.87} & 1.57 & 1.68 &       0.91 &       0.88 &         1.14 & 0.97 &     2.24 \\
    & WRN 40x10         &          3.12 & \textbf{1.16} & 1.23 & 2.30 &       1.29 &       1.32 &         1.39 & 1.29 &     3.49 \\
    \cline{2-11}
    \\[-1em]
    & Avg. Relative $\overline{ECE}$ &          1.00 & 0.42 & 0.57 & 0.66 & 0.42 & 0.40 & 0.44 & \textbf{0.39} & 1.04 \\
    \midrule
    \multirow{ 9}{*}{\rotatebox[origin=c]{90}{CIFAR 100}} 
    & DenseNet 121      &          8.76 & 3.85 & \textbf{3.01} & 3.78 &       3.20 &       5.25 &         4.76 & 3.21 &     5.67 \\
    & DenseNet 169      &          8.93 & 4.07 & \textbf{2.97} & 3.82 &       3.31 &       5.31 &         5.04 & 3.24 &     4.78 \\
    & ResNet 101        &         11.45 & 2.28 & 2.26 & 3.32 &       2.47 &       4.03 &         4.20 & \textbf{2.09} &     4.08 \\
    & ResNext 29 8x16   &          9.69 & 3.20 & 2.80 & 3.24 &       2.26 &       4.82 &         4.48 & \textbf{2.06} &     3.13 \\
    & VGG 19            &         17.63 & 5.08 & 5.25 & 4.20 &       4.02 &       4.00 &         4.61 & \textbf{3.96} &     6.14 \\
    & WRN 28x10         &          5.19 & 4.69 & 3.72 & 3.79 &       3.59 &       5.26 &         4.56 & \textbf{3.38} &     6.16 \\
    & WRN 40x10         &         14.78 & 4.34 & \textbf{3.16} & 4.46 &       4.09 &       5.46 &         5.00 & 4.21 &     7.89 \\
    \cline{2-11}
    \\[-1em]
    & Avg. Relative $\overline{ECE}$ &          1.00 & 0.42 & 0.34 & 0.39 & 0.34 & 0.52 & 0.49 & \textbf{0.33} & 0.56 \\
    \bottomrule
\end{tabular}}
\end{table}

\begin{table}
    \centering
    \caption{NLL using $1000$ validation samples. Models are denoted by their architecture and depth (and width if applicable).}
    \label{tab:nll_md}
    \resizebox{\textwidth}{!}{\begin{tabular}{llccccccccc}
    \toprule
    &  Task  &  Uncalibrated &   TS &  ETS  &  BTS &  HTS & LTS &  HnLTS & PTS
    & \begin{tabular}{@{}c@{}}PTS \\ ($L_{ECE}$)\end{tabular}  \\
    \midrule
    \multirow{ 10}{*}{\rotatebox[origin=c]{90}{CIFAR 10}} 
    & DenseNet 121      &          0.1881 & 0.1621 & 0.1685 & 0.3142 &     \textbf{0.1618} &     0.1651 &       0.1710 & 0.1648 &   0.2331 \\
    & DenseNet 169      &          0.1870 & 0.1611 & 0.1685 & 0.2973 &     0.1598 &     0.1649 &       0.1710 & \textbf{0.1594} &   0.2525 \\
    & ResNet 50         &          0.7897 & 0.4477 & 0.4520 & 0.5999 &     0.4461 &     0.4441 &       \textbf{0.4438} & 0.4476 &   0.7613 \\
    & ResNet 101        &          0.3047 & \textbf{0.2166} & 0.2199 & 0.4069 &     0.2175 &     0.2188 &       0.2217 & 0.2200 &   0.3255 \\
    & ResNext 29 8x16   &          0.1997 & \textbf{0.1624} & 0.1633 & 0.3835 &     0.1635 &     0.1719 &       0.1775 & 0.1730 &   0.2420 \\
    & VGG 19            &          0.2998 & 0.2358 & 0.2421 & 0.4146 &     \textbf{0.2337} &     0.2371 &       0.2426 & 0.2352 &   0.3991 \\
    & WRN 28x10         &          0.1497 & \textbf{0.1364} & 0.1419 & 0.2575 &     0.1367 &     0.1389 &       0.1491 & 0.1380 &   0.1730 \\
    & WRN 40x10         &          0.2068 & \textbf{0.1637} & 0.1653 & 0.3747 &     0.1642 &     0.1701 &       0.1730 & 0.1690 &   0.2813 \\
    \cline{2-11}
    \\[-1em]
    & Avg. Relative $\overline{NLL}$ &          1.0000 & 0.7880 & 0.8080 & 1.5240 & \textbf{0.7874} & 0.8053 & 0.8304 & 0.8010 & 1.2101 \\
    \midrule
    \multirow{ 9}{*}{\rotatebox[origin=c]{90}{CIFAR 100}} 
    & DenseNet 121      &          0.8939 & 0.8359 & 0.8601 & 0.9227 &     0.8260 &     0.9021 &       0.8849 & \textbf{0.8161} &   0.9383 \\
    & DenseNet 169      &          0.8748 & 0.8161 & 0.8326 & 0.9240 &     0.8070 &     0.8797 &       0.8682 & \textbf{0.7971} &   0.8577 \\
    & ResNet 101        &          1.1343 & 1.0010 & 1.0084 & 1.0979 &     1.0012 &     1.0636 &       1.0663 & \textbf{0.9992} &   1.0385 \\
    & ResNext 29 8x16   &          0.9398 & 0.8224 & 0.8407 & 0.9218 &     0.8133 &     0.9011 &       0.8938 & \textbf{0.8072} &   0.8238 \\
    & VGG 19            &          1.5414 & 1.1998 & 1.2064 & 1.2993 &     1.1947 &     1.2158 &       1.2194 & \textbf{1.1799} &   1.2454 \\
    & WRN 28x10         &          0.8173 & 0.8138 & 0.8334 & 0.9257 &     0.7873 &     0.8600 &       0.8242 & \textbf{0.7736} &   0.8644 \\
    & WRN 40x10         &          1.2248 & 0.9058 & 0.9259 & 0.9883 &     0.8997 &     0.9533 &       0.9309 & \textbf{0.8843} &   1.1977 \\
    \cline{2-11}
    \\[-1em]
    & Avg. Relative $\overline{NLL}$ &          1.0000 & 0.8770 & 0.8937 & 0.9742 & 0.8668 & 0.9329 & 0.9190 & \textbf{0.8568} & 0.9522 \\
    \bottomrule
\end{tabular}}
\end{table}

\begin{table}
    \centering
    \caption{Brier Score using $1000$ validation samples. Models are denoted by their architecture and depth (and width if applicable).}
    \label{tab:bri_md}
    \resizebox{\textwidth}{!}{\begin{tabular}{llccccccccc}
    \toprule
    &  Task  &  Uncalibrated &   TS &  ETS &  BTS &  HTS & LTS &  HnLTS & PTS & \begin{tabular}{@{}c@{}}PTS \\ ($L_{ECE}$)\end{tabular}  \\
    \midrule
    \multirow{10}{*}{\rotatebox[origin=c]{90}{CIFAR 10}} 
    & DenseNet 121      &          0.0764 & 0.0729 & 0.0726 & 0.0755 &     0.0727 &     0.0730 &       0.0732 & \textbf{0.0725} &   0.0782 \\
    & DenseNet 169      &          0.0754 & 0.0718 & 0.0713 & 0.0735 &     0.0714 &     0.0719 &       0.0718 & \textbf{0.0712} &   0.0777 \\
    & ResNet 50         &          0.2392 & 0.2037 & 0.2030 & 0.2110 &     0.2034 &     0.2030 &       \textbf{0.2026} & 0.2032 &   0.2281 \\
    & ResNet 101        &          0.1102 & 0.1011 & \textbf{0.1011} & 0.1069 &     0.1012 &     0.1011 &       0.1012 & 0.1011 &   0.1101 \\
    & ResNext 29 8x16   &          0.0828 & \textbf{0.0784} & 0.0784 & 0.0838 &     0.0786 &     0.0793 &       0.0796 & 0.0790 &   0.0853 \\
    & VGG 19            &          0.1101 & 0.1019 & 0.1008 & 0.1053 &     0.1009 &     0.1017 &       0.1011 & \textbf{0.1007} &   0.1131 \\
    & WRN 28x10         &          0.0629 & \textbf{0.0608} & 0.0610 & 0.0638 &     0.0609 &     0.0609 &       0.0614 & 0.0609 &   0.0644 \\
    & WRN 40x10         &          0.0820 & 0.0768 & \textbf{0.0766} & 0.0823 &     0.0770 &     0.0774 &       0.0775 & 0.0771 &   0.0851 \\
    \cline{2-11}
    \\[-1em]
    & Avg. Relative $\overline{Brier}$ &          1.0000 & 0.9317 & \textbf{0.9286} & 0.9754 & 0.9303 & 0.9340 & 0.9347 & 0.9300 & 1.0162 \\
    \midrule
    \multirow{ 9}{*}{\rotatebox[origin=c]{90}{CIFAR 100}}  
    & DenseNet 121      &          0.3171 & 0.3048 & 0.3046 & 0.3095 &     0.3047 &     0.3130 &       0.3126 & \textbf{0.3024} &   0.3117 \\
    & DenseNet 169      &          0.3142 & 0.3017 & 0.3013 & 0.3054 &     0.3014 &     0.3087 &       0.3085 & \textbf{0.2986} &   0.3055 \\
    & ResNet 101        &          0.4053 & 0.3818 & 0.3818 & 0.3874 &     0.3820 &     0.3908 &       0.3908 & \textbf{0.3810} &   0.3867 \\
    & ResNext 29 8x16   &          0.3275 & 0.3096 & 0.3095 & 0.3142 &     0.3093 &     0.3197 &       0.3201 & \textbf{0.3072} &   0.3096 \\
    & VGG 19            &          0.4433 & 0.3920 & 0.3911 & 0.3917 &     0.3904 &     0.3950 &       0.3943 & \textbf{0.3865} &   0.3946 \\
    & WRN 28x10         &          0.2892 & 0.2887 & 0.2880 & 0.2905 &     0.2871 &     0.2919 &       0.2900 & \textbf{0.2836} &   0.2935 \\
    & WRN 40x10         &          0.3700 & 0.3278 & 0.3273 & 0.3345 &     0.3295 &     0.3317 &       0.3312 & \textbf{0.3250} &   0.3475 \\
    \cline{2-11}
    \\[-1em]
    & Avg. Relative $\overline{Brier}$ &          1.0000 & 0.9397 & 0.9384 & 0.9508 & 0.9387 & 0.9581 & 0.9567 & \textbf{0.9304} & 0.9570 \\
    \bottomrule
\end{tabular}}
\end{table}

\begin{table}
    \centering
    \caption{ECE ($M=50$) using $500$ validation samples. Models are denoted by their architecture and depth (and width if applicable).}
    \label{tab:ece_mld}
    \resizebox{\textwidth}{!}{\begin{tabular}{llccccccccc}
    \toprule
    &  Task  &  Uncalibrated &   TS &  ETS  &  BTS &  HTS & LTS &  HnLTS & PTS & \begin{tabular}{@{}c@{}}PTS \\ ($L_{ECE}$)\end{tabular}  \\
    \midrule
    \multirow{10}{*}{\rotatebox[origin=c]{90}{CIFAR 10}} 
    & DenseNet 121      &          2.84 & 1.55 & 2.36 & 2.94 &       1.54 &       \textbf{1.33} &         1.76 & 1.90 &     2.99 \\
    & DenseNet 169      &          2.79 & 1.59 & 2.41 & 2.58 &       1.44 &       \textbf{1.32} &         1.66 & 1.87 &     3.14 \\
    & ResNet 50         &         10.71 & 2.69 & 2.76 & 5.09 &       \textbf{2.48} &       2.56 &         2.58 & 2.53 &     9.71 \\
    & ResNet 101        &          4.45 & \textbf{1.53} & 1.82  & 3.96 &       1.75 &       1.77 &         2.04 & 2.18 &     4.42 \\
    & ResNext 29 8x16   &          2.88 & \textbf{1.21} & 1.22 & 3.29 &       1.42 &       1.55 &         1.89 & 2.23 &     3.35 \\
    & VGG 19            &          4.61 & 2.58 & 3.29 & 3.88 &       \textbf{2.06} &       2.07 &         2.28 & 2.60 &     4.65 \\
    & WRN 28x10         &          1.93 & \textbf{0.90} & 1.80 & 2.37 &       0.97 &       0.95 &         1.40 & 1.62 &     2.37 \\
    & WRN 40x10         &          3.12 & \textbf{1.13} & 1.55 & 3.23 &       1.28 &       1.34 &         1.58 & 2.03 &     3.48 \\
    \cline{2-11}
    \\[-1em]
    & Avg. Relative $\overline{ECE}$ &          1.00 & 0.44 & 0.62 & 0.95 & 0.44 & \textbf{0.44} & 0.54 & 0.61 & 1.07 \\
    \midrule
    \multirow{ 9}{*}{\rotatebox[origin=c]{90}{CIFAR 100}} 
    & DenseNet 121      &          8.76 & 4.04 & \textbf{3.30} & 4.94 &       3.44 &       7.55 &         8.12 & 3.54 &     8.26 \\
    & DenseNet 169      &          8.93 & 4.09 & \textbf{3.19} & 4.89 &       3.41 &       7.62 &         8.59 & 3.54 &     8.31 \\
    & ResNet 101        &         11.45 & 2.29 & 2.48 & 4.62 &       2.51 &       6.31 &         7.24 & \textbf{2.23} &     5.17 \\
    & ResNext 29 8x16   &          9.69 & 3.34 & 2.99 & 4.99 &       \textbf{2.52} &       7.62 &         8.86 & 2.53 &     6.31 \\
    & VGG 19            &         17.63 & 5.08 & 5.33 & 5.27 &       4.07 &       5.56 &         6.50 & \textbf{3.82} &     6.62 \\
    & WRN 28x10         &          5.19 & 4.73 & 3.80 & 4.77 &       3.63 &       6.65 &         7.00 & \textbf{3.54} &     6.78 \\
    & WRN 40x10         &         14.78 & 4.12 & \textbf{3.33} & 5.30 &       3.81 &       6.88 &         6.67 & 4.03 &     9.67 \\
    \cline{2-11}
    \\[-1em]
    & Avg. Relative $\overline{ECE}$ &          1.00 & 0.42 & 0.36 & 0.52 & 0.35 & 0.73 & 0.80 & \textbf{0.35} & 0.76 \\
    \bottomrule
\end{tabular}}
\end{table}

\begin{table}
    \centering
    \caption{NLL using $500$ validation samples. Models are denoted by their architecture and depth (and width if applicable).}
    \label{tab:nll_mld}
    \resizebox{\textwidth}{!}{\begin{tabular}{llccccccccc}
    \toprule
    &  Task  &  Uncalibrated &   TS &  ETS  &  BTS &  HTS & LTS &  HnLTS & PTS
    & \begin{tabular}{@{}c@{}}PTS \\ ($L_{ECE}$)\end{tabular}  \\
    \midrule
    \multirow{ 10}{*}{\rotatebox[origin=c]{90}{CIFAR 10}} 
    & DenseNet 121      &          0.1881 & \textbf{0.1623} & 0.1690 & 0.4336 &     0.1625 &     0.1765 &       0.2032 & 0.2051 &   0.2180 \\
    & DenseNet 169      &          0.1870 & \textbf{0.1617} & 0.1686 & 0.4080 &     0.1623 &     0.1728 &       0.1969 & 0.2225 &   0.2386 \\
    & ResNet 50         &          0.7897 & 0.4481 & 0.4515 & 0.9523 &     \textbf{0.4474} &     0.4529 &       0.4545 & 0.4740 &   0.9776 \\
    & ResNet 101        &          0.3047 & \textbf{0.2170} & 0.2212 & 0.5993 &     0.2184 &     0.2310 &       0.2476 & 0.2664 &   0.3541 \\
    & ResNext 29 8x16   &          0.1997 & \textbf{0.1640} & 0.1646 & 0.5475 &     0.1663 &     0.1895 &       0.2304 & 0.2754 &   0.2628 \\
    & VGG 19            &          0.2998 & 0.2362 & 0.2433 & 0.5869 &     \textbf{0.2348} &     0.2468 &       0.2825 & 0.2895 &   0.3802 \\
    & WRN 28x10         &          0.1497 & \textbf{0.1367} & 0.1438 & 0.3427 &     0.1378 &     0.1463 &       0.1757 & 0.1835 &   0.1813 \\
    & WRN 40x10         &          0.2068 & \textbf{0.1638} & 0.1679 & 0.5126 &     0.1651 &     0.1803 &       0.1989 & 0.2361 &   0.2742 \\
    \cline{2-11}
    \\[-1em]
    & Avg. Relative $\overline{NLL}$ &          1.0000 & \textbf{0.7902} & 0.8132  & 2.1410 & 0.7938 & 0.8519 & 0.9691 & 1.0584 & 1.2445 \\
    \midrule
    \multirow{ 9}{*}{\rotatebox[origin=c]{90}{CIFAR 10}} 
    & DenseNet 121      &          0.8939 & 0.8367 & 0.8604 & 1.0944 &     0.8268 &     1.0817 &       1.1304 & \textbf{0.8238} &   0.9738 \\
    & DenseNet 169      &          0.8748 & 0.8164 & 0.8346 & 1.0912 &     0.8075 &     1.0688 &       1.1528 & \textbf{0.8040} &   1.0126 \\
    & ResNet 101        &          1.1343 & \textbf{1.0012} & 1.0117 & 1.2453 &     1.0017 &     1.2018 &       1.2464 & 1.0022 &   1.0918 \\
    & ResNext 29 8x16   &          0.9398 & 0.8234 & 0.8410 & 1.1430 &     0.8146 &     1.1242 &       1.2370 & \textbf{0.8124} &   0.9120 \\
    & VGG 19            &          1.5414 & 1.2004 & 1.2079 & 1.4489 &     1.1955 &     1.2952 &       1.3199 & \textbf{1.1796} &   1.2775 \\
    & WRN 28x10         &          0.8173 & 0.8147 & 0.8335 & 1.0770 &     0.7883 &     0.9975 &       1.0226 & \textbf{0.7791} &   0.8900 \\
    & WRN 40x10         &          1.2248 & 0.9065 & 0.9315 & 1.2020 &     0.9007 &     1.0961 &       1.0846 & \textbf{0.8899} &   1.2474 \\
    \cline{2-11}
    \\[-1em]
    & Avg. Relative $\overline{NLL}$ &          1.0000 & 0.8777 & 0.8953  & 1.1464 & 0.8676 & 1.0919 & 1.1415 & \textbf{0.8619} & 1.0166 \\
    \bottomrule
\end{tabular}}
\end{table}

\begin{table}
    \centering
    \caption{Brier Score using $500$ validation samples. Models are denoted by their architecture and depth (and width if applicable).}
    \label{tab:bri_mld}
    \resizebox{\textwidth}{!}{\begin{tabular}{llccccccccc}
    \toprule
    &  Task  &  Uncalibrated &   TS &  ETS &  BTS &  HTS & LTS &  HnLTS & PTS & \begin{tabular}{@{}c@{}}PTS \\ ($L_{ECE}$)\end{tabular}  \\
    \midrule
    \multirow{ 10}{*}{\rotatebox[origin=c]{90}{CIFAR 10}} 
    & DenseNet 121      &          0.0764 & 0.0729 & \textbf{0.0727} & 0.0790 &     0.0728 &     0.0736 &       0.0745 & 0.0751 &   0.0778 \\
    & DenseNet 169      &          0.0754 & 0.0718 & \textbf{0.0713} & 0.0770 &     0.0715 &     0.0724 &       0.0729 & 0.0738 &   0.0777 \\
    & ResNet 50         &          0.2392 & 0.2040 & \textbf{0.2032} & 0.2228 &     0.2038 &     0.2045 &       0.2043 & 0.2061 &   0.2394 \\
    & ResNet 101        &          0.1102 & \textbf{0.1012} & 0.1013 & 0.1135 &     0.1014 &     0.1022 &       0.1027 & 0.1042 &   0.1114 \\
    & ResNext 29 8x16   &          0.0828 & \textbf{0.0786} & 0.0786 & 0.0895 &     0.0788 &     0.0807 &       0.0821 & 0.0838 &   0.0862 \\
    & VGG 19            &          0.1101 & 0.1020 & \textbf{0.1009} & 0.1110 &     0.1010 &     0.1026 &       0.1027 & 0.1043 &   0.1118 \\
    & WRN 28x10         &          0.0629 & \textbf{0.0608} & 0.0611 & 0.0668 &     0.0609 &     0.0615 &       0.0627 & 0.0637 &   0.0649 \\
    & WRN 40x10         &          0.0820 & \textbf{0.0768} & 0.0769 & 0.0878 &     0.0769 &     0.0781 &       0.0787 & 0.0814 &   0.0849 \\
    \cline{2-11}
    \\[-1em]
    & Avg. Relative $\overline{Brier}$ &          1.0000 & 0.9323 & \textbf{0.9303} & 1.0301 & 0.9314 & 0.9429 & 0.9511 & 0.9671 & 1.0232 \\
    \midrule
    \multirow{ 9}{*}{\rotatebox[origin=c]{90}{CIFAR 100}} 
    & DenseNet 121      &          0.3171 & 0.3051 & 0.3051 & 0.3151 &     0.3051 &     0.3294 &       0.3313 & \textbf{0.3040} &   0.3209 \\
    & DenseNet 169      &          0.3142 & 0.3018 & 0.3015 & 0.3109 &     0.3016 &     0.3259 &       0.3295 & \textbf{0.2997} &   0.3202 \\
    & ResNet 101        &          0.4053 & 0.3818 & 0.3820 & 0.3929 &     0.3820 &     0.4056 &       0.4071 & \textbf{0.3816} &   0.3923 \\
    & ResNext 29 8x16   &          0.3275 & 0.3098 & 0.3099 & 0.3219 &     0.3095 &     0.3387 &       0.3433 & \textbf{0.3081} &   0.3204 \\
    & VGG 19            &          0.4433 & 0.3920 & 0.3913 & 0.3958 &     0.3903 &     0.4051 &       0.4043 & \textbf{0.3861} &   0.3966 \\
    & WRN 28x10         &          0.2892 & 0.2888 & 0.2882 & 0.2952 &     0.2872 &     0.3039 &       0.3048 & \textbf{0.2843} &   0.2960 \\
    & WRN 40x10         &          0.3700 & 0.3277 & 0.3276 & 0.3421 &     0.3295 &     0.3455 &       0.3455 & \textbf{0.3255} &   0.3547 \\
    \cline{2-11}
    \\[-1em]
    & Avg. Relative $\overline{Brier}$ &          1.0000 & 0.9399 & 0.9393 & 0.9677 & 0.9391 & 1.0013 & 1.0066 & \textbf{0.9327} & 0.9792 \\
    \bottomrule
\end{tabular}}
\end{table}


\begin{table}
    \centering
    \caption{ECE ($M=50$) using $200$ validation samples. Models are denoted by their architecture and depth (and width if applicable).}
    \label{tab:ece_ld}
    \resizebox{\textwidth}{!}{\begin{tabular}{llccccccccc}
    \toprule
     &  Task  &  Uncalibrated &   TS &  ETS  &  BTS &  HTS & LTS &  HnLTS & PTS & \begin{tabular}{@{}c@{}}PTS \\ ($L_{ECE}$)\end{tabular}  \\
    \midrule
    \multirow{ 10}{*}{\rotatebox[origin=c]{90}{CIFAR 10}} 
    & DenseNet 121      &          2.84 & \textbf{1.72} & 2.24 & 4.01 &       1.77 &       1.96 &         3.00 & 3.69 &     3.08 \\
    & DenseNet 169      &          2.79 & 1.70 & 2.86 & 3.99 &       \textbf{1.56} &       1.74 &         2.66 & 3.62 &     3.05 \\
    & ResNet 50         &         10.71 & 2.69 & 3.23 & 7.88 &       \textbf{2.51} &       2.92 &         3.19 & 5.47 &    10.04 \\
    & ResNet 101        &          4.45 & \textbf{1.81} & 2.16 & 5.11 &       2.03 &       2.26 &         3.11 & 4.45 &     4.67 \\
    & ResNext 29 8x16   &          2.88 & \textbf{1.34} & 1.72 & 4.12 &       1.51 &       1.87 &         3.00 & 3.95 &     3.04 \\
    & VGG 19            &          4.61 & 2.59 & 3.16 & 5.42 &       \textbf{2.40} &       2.83 &         3.77 & 4.96 &     4.96 \\
    & WRN 28x10         &          1.93 & \textbf{1.10} & 2.24 & 3.56 &       1.23 &       1.46 &         2.46 & 2.73 &     2.43 \\
    & WRN 40x10         &          3.12 & \textbf{1.39} & 1.89 & 4.08 &       1.62 &       1.92 &         2.96 & 3.77 &     3.37 \\
    \cline{2-11}
    \\[-1em]
    & Avg. Relative $\overline{ECE}$ &          1.00 & \textbf{0.49} & 0.71 & 1.31 & 0.51 & 0.59 & 0.89 & 1.15 & 1.08 \\
    \midrule
    \multirow{ 9}{*}{\rotatebox[origin=c]{90}{CIFAR 100}} 
    & DenseNet 121      &          8.76 & 4.01 & \textbf{3.44} & 8.78 &       3.49 &      19.00 &        17.72 & 4.77 &     9.01 \\
    & DenseNet 169      &          8.93 & 4.22 & \textbf{3.44} & 8.56 &       3.68 &      18.93 &        17.19 & 5.19 &     9.76 \\
    & ResNet 101        &         11.45 & \textbf{2.77} & 3.17 & 8.59 &       3.11 &      17.43 &        19.18 & 3.98 &     9.94 \\
    & ResNext 29 8x16   &          9.69 & 3.39 & 3.23 & 8.55 &       \textbf{2.73} &      19.13 &        18.16 & 3.55 &     7.13 \\
    & VGG 19            &         17.63 & 5.16 & 5.53 & 8.46 &       \textbf{4.41} &      14.55 &        15.33 & 4.48 &    11.71 \\
    & WRN 28x10         &          5.19 & 4.88 & 4.18 & 7.93 &       \textbf{4.01} &      13.85 &        15.35 & 5.34 &     8.12 \\
    & WRN 40x10         &         14.78 & 4.52 & \textbf{3.83} & 9.64 &       4.47 &      15.71 &        17.47 & 5.23 &     9.87 \\
    \cline{2-11}
    \\[-1em]
    & Avg. Relative $\overline{ECE}$ &          1.00 & 0.44 & 0.40 & 0.89 & \textbf{0.38} & 1.76 & 1.79 & 0.50 & 0.95 \\
    \bottomrule
\end{tabular}}
\end{table}

\begin{table}
    \centering
    \caption{NLL using $200$ validation samples. Models are denoted by their architecture and depth (and width if applicable).}
    \label{tab:nll_ld}
    \resizebox{\textwidth}{!}{\begin{tabular}{llccccccccc}
    \toprule
    &  Model  &  Uncalibrated &   TS &  ETS &  BTS &  HTS & LTS &  HnLTS & PTS
    & \begin{tabular}{@{}c@{}}PTS \\ ($L_{ECE}$)\end{tabular}  \\
    \midrule
    \multirow{ 10}{*}{\rotatebox[origin=c]{90}{CIFAR 10}} 
    & DenseNet 121      &          0.1881 & \textbf{0.1654} & 0.1709 & 0.6331 &     0.1671 &     0.2511 &       0.3918 & 0.4297 &   0.2263 \\
    & DenseNet 169      &          0.1870 & \textbf{0.1633} & 0.1748 & 0.5877 &     0.1646 &     0.2285 &       0.3335 & 0.4008 &   0.2342 \\
    & ResNet 50         &          0.7897 & \textbf{0.4489} & 0.4553 & 1.4653 &     0.4489 &     0.5041 &       0.5241 &    $\infty$ &   0.8810 \\
    & ResNet 101        &          0.3047 & \textbf{0.2199} & 0.2248 & 0.7817 &     0.2256 &     0.2811 &       0.4023 & 0.5189 &   0.3902 \\
    & ResNext 29 8x16   &          0.1997 & \textbf{0.1668} & 0.1734 & 0.6465 &     0.1698 &     0.2523 &       0.3968 & 0.4850 &   0.2282 \\
    & VGG 19            &          0.2998 & \textbf{0.2377} & 0.2422 & 0.8606 &     0.2385 &     0.3304 &       0.4858 & 0.5694 &   0.3981 \\
    & WRN 28x10         &          0.1497 & \textbf{0.1381} & 0.1476 & 0.4952 &     0.1457 &     0.1955 &       0.2971 & 0.2930 &   0.1830 \\
    & WRN 40x10         &          0.2068 & \textbf{0.1662} & 0.1746 & 0.6286 &     0.1766 &     0.2525 &       0.3927 & 0.4728 &   0.2533 \\
    \cline{2-11}
    \\[-1em]
    & Avg. Relative $\overline{NLL}$ &          1.0000 & \textbf{0.7997} & 0.8329 & 2.9233 & 0.8188 & 0.1263 & 1.6678 & 2.1004 & 1.2213 \\
    \midrule
    \multirow{ 9}{*}{\rotatebox[origin=c]{90}{CIFAR 100}} 
    & DenseNet 121      &          0.8939 & 0.8374 & 0.8595 & 1.7011 &     \textbf{0.8285} &     2.2353 &       2.3776 & 0.9091 &   1.0147 \\
    & DenseNet 169      &          0.8748 & 0.8181 & 0.8359 & 1.6978 &     \textbf{0.8089} &     2.1935 &       2.2927 & 0.9039 &   0.9845 \\
    & ResNet 101        &          1.1343 & \textbf{1.0036} & 1.0147 & 1.8572 &     1.0059 &     2.3447 &       2.7525 & 1.0721 &   1.2336 \\
    & ResNext 29 8x16   &          0.9398 & 0.8248 & 0.8420 & 1.6495 &     \textbf{0.8174} &     2.3144 &       2.4890 & 0.8691 &   0.9635 \\
    & VGG 19            &          1.5414 & 1.2013 & 1.2099 & 1.8499 &     \textbf{1.1977} &     2.0490 &       2.2260 & 1.2177 &   1.6701 \\
    & WRN 28x10         &          0.8173 & 0.8168 & 0.8362 & 1.5552 &     \textbf{0.7920} &     1.6668 &       2.0215 & 0.9093 &   0.9380 \\
    & WRN 40x10         &          1.2248 & 0.9073 & 0.9323 & 1.9061 &     \textbf{0.9046} &     2.0570 &       2.4555 & 0.9438 &   1.2121 \\
    \cline{2-11}
    \\[-1em]
    & Avg. Relative $\overline{NLL}$ &          1.0000 & 0.8791 & 0.8967 & 1.6993 & \textbf{0.8704} & 2.0837 & 2.3254 & 0.9419 & 1.0849 \\
    \bottomrule
\end{tabular}}
\end{table}

\begin{table}
    \centering
    \caption{Brier Score using $200$ validation samples. Models are denoted by their architecture and depth (and width if applicable).}
    \label{tab:bri_ld}
    \resizebox{\textwidth}{!}{\begin{tabular}{llccccccccc}
    \toprule
      &  Task  &  Uncalibrated &   TS &  ETS &  BTS &  HTS & LTS &  HnLTS & PTS & \begin{tabular}{@{}c@{}}PTS \\ ($L_{ECE}$)\end{tabular}  \\
    \midrule
    \multirow{10}{*}{\rotatebox[origin=c]{90}{CIFAR 10}} 
    & DenseNet 121      &          0.0764 & 0.0734 & \textbf{0.0733} & 0.0852 &     0.0734 &     0.0766 &       0.0805 & 0.0841 &   0.0781 \\
    & DenseNet 169      &          0.0754 & 0.0719 & 0.0720 & 0.0849 &     \textbf{0.0718} &     0.0746 &       0.0780 & 0.0833 &   0.0774 \\
    & ResNet 50         &          0.2392 & 0.2038 & 0.2039 & 0.2409 &     \textbf{0.2037} &     0.2092 &       0.2093 &    $\infty$ &   0.2379 \\
    & ResNet 101        &          0.1102 & \textbf{0.1016} & 0.1019 & 0.1200 &     0.1020 &     0.1051 &       0.1082 & 0.1161 &   0.1129 \\
    & ResNext 29 8x16   &          0.0828 & \textbf{0.0789} & 0.0801 & 0.0946 &     0.0792 &     0.0834 &       0.0884 & 0.0939 &   0.0844 \\
    & VGG 19            &          0.1101 & 0.1025 & \textbf{0.1012} & 0.1196 &     0.1019 &     0.1065 &       0.1099 & 0.1166 &   0.1133 \\
    & WRN 28x10         &          0.0629 & \textbf{0.0610} & 0.0617 & 0.0742 &     0.0611 &     0.0638 &       0.0675 & 0.0694 &   0.0651 \\
    & WRN 40x10         &          0.0820 & \textbf{0.0772} & 0.0781 & 0.0917 &     0.0776 &     0.0816 &       0.0859 & 0.0906 &   0.0842 \\
    \cline{2-11}
    \\[-1em]
    & Avg. Relative $\overline{Brier}$ &          1.0000 & \textbf{0.9357} & 0.9393 & 1.1080 & 0.9363 & 0.9758 & 1.0167 & 1.0946 & 1.0225 \\
    \midrule
    \multirow{ 9}{*}{\rotatebox[origin=c]{90}{CIFAR 100}} 
    & DenseNet 121      &          0.3171 & 0.3052 & 0.3053 & 0.3355 &     \textbf{0.3049} &     0.4109 &       0.3948 & 0.3118 &   0.3238 \\
    & DenseNet 169      &          0.3142 & 0.3022 & \textbf{0.3019} & 0.3306 &     0.3020 &     0.4073 &       0.3876 & 0.3098 &   0.3224 \\
    & ResNet 101        &          0.4053 & \textbf{0.3824} & 0.3830 & 0.4161 &     0.3827 &     0.4813 &       0.4842 & 0.3889 &   0.4096 \\
    & ResNext 29 8x16   &          0.3275 & 0.3101 & 0.3103 & 0.3402 &     \textbf{0.3097} &     0.4215 &       0.4079 & 0.3130 &   0.3244 \\
    & VGG 19            &          0.4433 & 0.3921 & 0.3917 & 0.4118 &     \textbf{0.3907} &     0.4588 &       0.4556 & 0.3911 &   0.4246 \\
    & WRN 28x10         &          0.2892 & 0.2892 & 0.2891 & 0.3120 &     \textbf{0.2877} &     0.3557 &       0.3583 & 0.2954 &   0.2996 \\
    & WRN 40x10         &          0.3700 & \textbf{0.3282} & 0.3284 & 0.3667 &     0.3306 &     0.4107 &       0.4160 & 0.3327 &   0.3544 \\
    \cline{2-11}
    \\[-1em]
    & Avg. Relative $\overline{Brier}$ &          1.0000 & 0.9409 & 0.9410 & 1.0250 & \textbf{0.9404} & 1.2059 & 1.1872 & 0.9554 & 1.0000 \\
    \bottomrule
\end{tabular}}
\end{table}

\clearpage

 \bibliographystyle{elsarticle-num-names}





\end{document}